\PassOptionsToPackage{table, dvipsnames}{xcolor}
\documentclass[sigconf]{acmart}
\AtBeginDocument{%
  }

\usepackage{booktabs}
\usepackage{array}
\usepackage{balance} 
\usepackage{multirow}
\usepackage[normalem]{ulem}
\usepackage{color}
\definecolor{lightgray}{RGB}{215,215,215}
\usepackage{colortbl}
\useunder{\uline}{\myul}{}
\usepackage{subfigure}
\usepackage{algorithm}  
\usepackage{algorithmicx}  
\usepackage[noend]{algpseudocode}  
\usepackage{stackengine}

\usepackage{amsmath}  
\usepackage{enumitem}
\usepackage{tabularx}
\usepackage[utf8]{inputenc}
\usepackage[english]{babel}
\usepackage{amsthm}
\usepackage{bm}
\usepackage{threeparttable}

\usepackage[most]{tcolorbox}
\usepackage{graphicx}
\definecolor{style}{RGB}{214,223,237}
\definecolor{semantics}{RGB}{212,233,226}

\usepackage{adjustbox}
\newcommand{\style}[1]{\adjustbox{valign=c,bgcolor=style}{#1}}
\newcommand{\semantics}[1]{\adjustbox{valign=c,bgcolor=semantics}{#1}}

\newcommand{\eg}{\emph{e.g., }}

\newlength\myindent
\usepackage{balance}

\usepackage{soul}

\floatname{algorithm}{Algorithm}

\copyrightyear{2025}
\acmYear{2025}
\setcopyright{acmlicensed}\acmConference[MM '25]{Proceedings of the 33rd ACM International Conference on Multimedia}{October 27--31, 2025}{Dublin, Ireland}
\acmBooktitle{Proceedings of the 33rd ACM International Conference on Multimedia (MM '25), October 27--31, 2025, Dublin, Ireland}
\acmDOI{10.1145/3746027.3755051}
\acmISBN{979-8-4007-2035-2/2025/10}

\begin{document}

\title{DRC: Enhancing Personalized Image Generation via Disentangled Representation Composition}

\author{Yiyan Xu}
\email{yiyanxu24@gmail.com}
\affiliation{
\institution{University of Science and Technology of China}
\country{China}
\city{Hefei}
}

\author{Wuqiang Zheng}
\email{qqqqqzheng@gmail.com}
\affiliation{
\institution{University of Science and Technology of China}
\country{China}
\city{Hefei}
}

\author{Wenjie Wang}
\email{wenjiewang96@gmail.com}
\affiliation{
\institution{University of Science and Technology of China}
\country{China}
\city{Hefei}
}

\author{Fengbin Zhu}
\authornote{Corresponding authors. This work is supported by the National Natural Science Foundation of China (62272437).}
\email{zhfengbin@gmail.com}
\affiliation{
\institution{National University of Singapore}
\country{Singapore}
\city{Singapore}
}

\author{Xinting Hu}
\authornotemark[1]
\email{xinting001@e.ntu.edu.sg}
\affiliation{
\institution{Nanyang Technological University}
\country{Singapore}
\city{Singapore}
}

\author{Yang Zhang}
\email{zyang1580@gmail.com}
\affiliation{
\institution{National University of Singapore}
\country{Singapore}
\city{Singapore}
}

\author{Fuli Feng}
\email{fulifeng93@gmail.com}
\affiliation{
\institution{University of Science and Technology of China}
\country{China}
\city{Hefei}
}

\author{Tat-Seng Chua}
\email{dcscts@nus.edu.sg}
\affiliation{
\institution{National University of Singapore}
\country{Singapore}
\city{Singapore}
}

\renewcommand{\shortauthors}{Yiyan Xu et al.}

\begin{abstract}
Personalized image generation has emerged as a promising direction in multimodal content creation. It aims to synthesize images tailored to individual \style{style preferences} (\eg color schemes, character appearances, layout) and \semantics{semantic intentions} (\eg emotion, action, scene contexts) by leveraging user-interacted history images and multimodal instructions. 
Despite notable progress, existing methods --- whether based on diffusion models, large language models, or Large Multimodal Models (LMMs) --- struggle to accurately capture and fuse user style preferences and semantic intentions. In particular, the state-of-the-art LMM-based method suffers from the entanglement of visual features, leading to \textit{Guidance Collapse}, where the generated images fail to preserve user-preferred styles or reflect the specified semantics. 

To address these limitations, we introduce \textbf{DRC}, a novel personalized image generation framework that enhances LMMs through \textit{\textbf{D}isentangled \textbf{R}epresentation \textbf{C}omposition}. DRC explicitly extracts user style preferences and semantic intentions from history images and the reference image, respectively, to form user-specific latent instructions that guide image generation within LMMs. Specifically, it involves two critical learning stages: \textbf{1) Disentanglement learning}, which employs a dual-tower disentangler to explicitly separate style and semantic features, optimized via a reconstruction-driven paradigm with difficulty-aware importance sampling; and  \textbf{2) Personalized modeling}, which applies semantic-preserving augmentations to effectively adapt the disentangled representations for robust personalized generation. Extensive experiments on two benchmarks demonstrate that DRC shows competitive performance while effectively mitigating the guidance collapse issue, underscoring the importance of disentangled representation learning for controllable and effective personalized image generation.

\end{abstract}




\maketitle

\vspace{-0.1cm}
\section{Introduction}
\label{sec:introduction}

\begin{figure}[t]
% \vspace{-0.2cm}
\setlength{\abovecaptionskip}{0.05cm}
\setlength{\belowcaptionskip}{-0.4cm}
\centering
\includegraphics[scale=0.65]{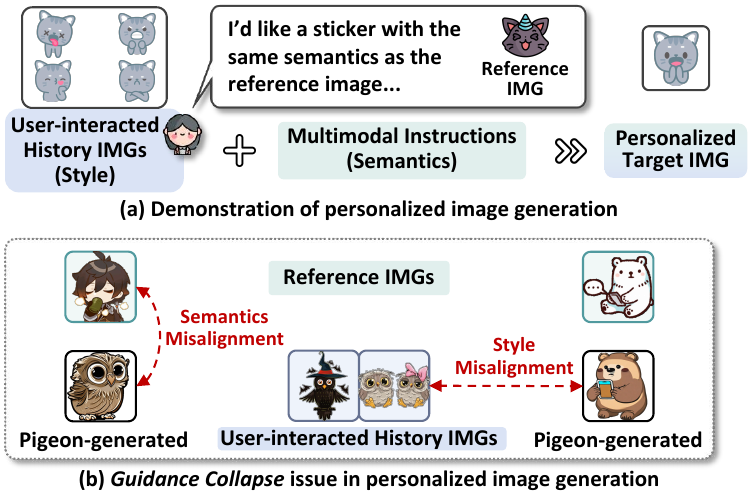}
\caption{Illustration of personalized image generation task and the \textit{Guidance Collapse} issue.} 
\label{fig:task}
\end{figure}

Recent advancements in large generative models have revolutionized multimodal content creation, driving a paradigm shift from generic to personalized generation~\cite{xu2025personalized}. Among these advances, personalized image generation, which aims to synthesize images tailored to specific users based on user-interacted history images and their multimodal instructions (see Figure~\ref{fig:task}(a)), has emerged as a particularly promising direction with broad application scenarios, such as personalized movie poster creation on streaming platforms like Netflix~\cite{xu2024personalized} and customized advertisement generation on e-commerce platforms like Amazon~\cite{yang2024new}. Recognizing its significant economic and research potential, this area has attracted increasing attention from both academia and industry~\cite{shen2024pmg,vashishtha2024chaining,xu2024diffusion}.

Existing studies on personalized image generation primarily build on Diffusion Models (DMs), Large Language Models (LLMs), or Large Multimodal Models (LMMs):
\begin{itemize}[leftmargin=*]
    \item \textbf{DM-based methods}~\cite{xu2024diffusion,wang2024g,xu2024sgdm} learn user preference representations from history images and integrate them with user instructions to guide the generation in DMs. However, they often struggle to interpret complex multimodal instructions precisely, leading to unsatisfactory performance~\cite{shen2024pmg,xu2024personalized}.

    \item \textbf{LLM-based methods}~\cite{shen2024pmg,chen2024tailored} leverage LLMs to capture user visual preferences and content needs from text by either converting images into textual descriptions or directly utilizing associated text prompts. While capable of handling complex instructions to some extent, these methods face challenges in extracting fine-grained visual details from history images and multimodal instructions, limiting their effectiveness in achieving precise visual personalization. 

    \item \textbf{LMM-based method}, Pigeon~\cite{xu2024personalized}, leverages LMMs to convert history images and multimodal instructions into visual and textual tokens, and then infer user preferences and requirements for generation, showcasing the superiority over previous methods. 
    Nevertheless, the entangled nature of style and semantic features in images makes it difficult to implicitly identify and utilize task-relevant cues to guide generation. As depicted in Figure~\ref{fig:task}(b), Pigeon fails to accurately capture the desired semantics specified by the reference image (left) or preserve the user-preferred style in history images (right), which we refer to as \textit{Guidance Collapse}.

\end{itemize}

\begin{figure}[t]
% \vspace{-0.2cm}
\setlength{\abovecaptionskip}{0.05cm}
\setlength{\belowcaptionskip}{-0.3cm}
\centering
\includegraphics[scale=0.55]{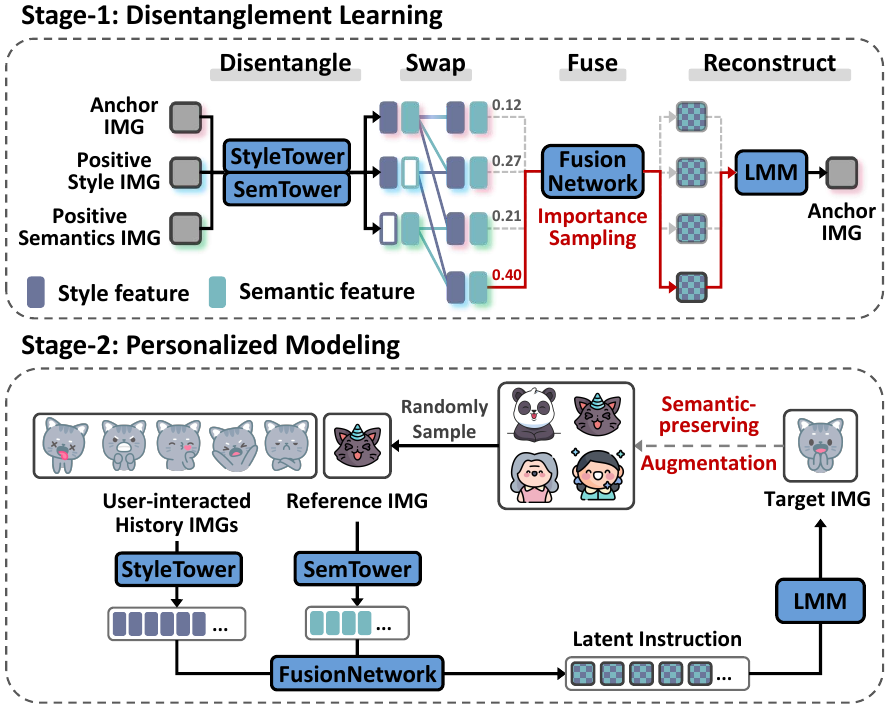}
\caption{Overview of DRC. A framework that first learns style-semantics disentanglement under a reconstruction-driven paradigm and then transfers this disentanglement capability for personalized image generation.} 
\label{fig:intro}
\end{figure}

Re-examining the personalized image generation task, the key lies in accurately capturing user \style{style preferences} (\eg color schemes, character appearances, and layout) from user-interacted history images and combining them with \semantics{semantic intentions} (\eg emotion, action, scene contexts) expressed in multimodal instructions. This motivates the exploration of an explicit style-semantics disentanglement mechanism to address the \textit{Guidance Collapse} issue. However, two major challenges persist: 1) Image style and semantics are globally entangled within high-dimensional feature space; the lack of direct supervision hampers effective disentanglement. 
2) Even after disentanglement, the extracted features often reside in a representation space that is not naturally aligned with the latent space in LMMs, thus limiting their utility for guiding LMMs' generation.

To tackle these challenges, we propose DRC, a framework to enhance LMM-based personalized image generation through \textit{\textbf{D}isentangled \textbf{R}epresentation \textbf{C}omposition}. 
As illustrated in Figure~\ref{fig:intro}, DRC consists of two critical stages: 
\begin{itemize}[leftmargin=*]
    \item \textbf{Stage-1: Disentanglement learning.} 
    DRC employs a dual-tower disentangler to extract image style and semantics separately, leveraging a reconstruction paradigm that naturally aligns well with LMM generation objectives to promote effective disentanglement. Specifically, we construct image triplets of <anchor image, positive-style image, positive-semantics image> for supervision, where DRC first extracts disentangled representations from the triplet images and systematically recombines them to create positive style-semantics pairings. These pairings are then integrated into latent instructions to guide anchor image reconstruction via LMMs, reinforcing the alignment between disentangled representation and the LMM latent space. Moreover, to improve training efficiency and disentanglement quality, we introduce a difficulty-aware importance sampling strategy to prioritize more challenging combinations during training.
    \item \textbf{Stage-2: Personalized modeling.} Building upon the well-disentangled representations learned in the first stage, DRC advances personalized image generation by composing style representations extracted from history images with semantic representations derived from the reference image, forming user-specific latent instructions to guide LMM-based image generation. To further enhance diversity and robustness, we introduce semantic-preserving augmentations to the target image, synthesizing multiple reference variants that maintain semantic consistency while varying in style. During training, we randomly sample one of these augmented references to encourage the model to extract invariant semantic representations and thereby improve the effectiveness of personalized image generation.
\end{itemize}

We conduct extensive experiments on two scenarios, following~\cite{xu2024personalized}: personalized sticker and movie poster generation, comparing DRC with various baselines. Quantitative and qualitative evaluations demonstrate that DRC shows competitive performance while achieving effective disentanglement to mitigate the guidance collapse issue. 
Our code and data are publicly available at \url{https://github.com/ZhengWwwq/DRC}.

In summary, our contributions can be concluded as follows: 

\begin{itemize}[leftmargin=*]
    \item We propose DRC, a novel framework that explicitly disentangles user style preferences and semantic intentions to compose user-specific latent instructions, pushing the boundaries of LMM-based personalized image generation by effectively mitigating guidance collapse.

    \item We design a two-stage learning strategy that first establishes effective style–semantic disentanglement for LMMs through a reconstruction-driven paradigm with difficulty-aware optimization, then utilizes semantic-preserving augmentations to transfer the disentanglement capability to personalized image generation.

    \item We conduct substantial experiments across two representative scenarios, providing strong empirical evidence of DRC's superiority over existing methods and offering new insights into the design of controllable, user-centric image generation models.

\end{itemize}

\section{DRC}

\begin{figure*}[t]
% \vspace{-0.2cm}
\setlength{\abovecaptionskip}{0.1cm}
\setlength{\belowcaptionskip}{-0.2cm}
\centering
\includegraphics[scale=0.69]{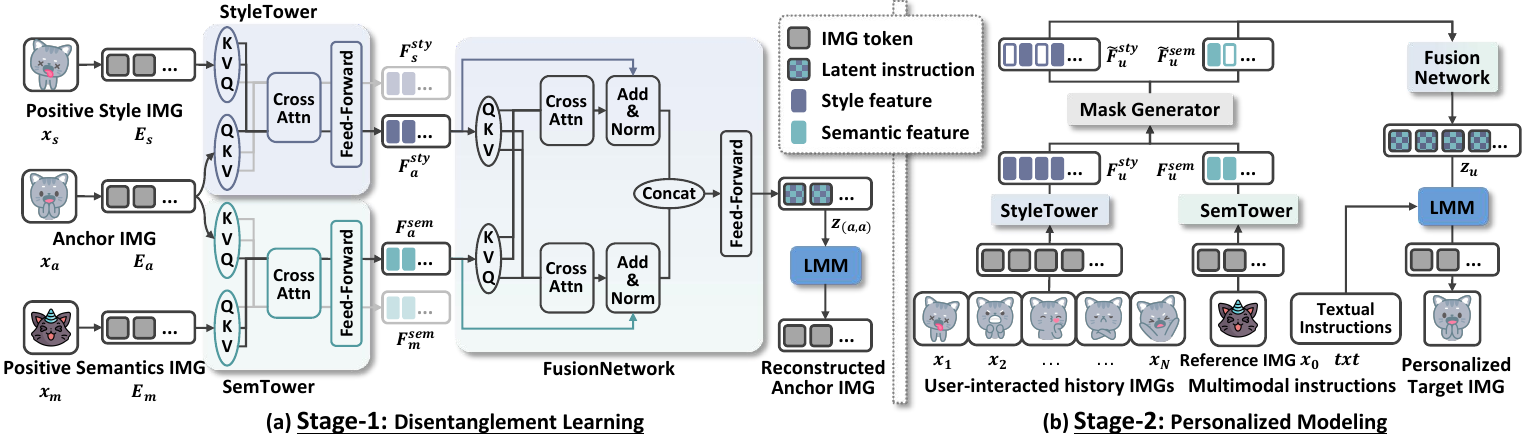}
\caption{Illustration of DRC, which consists of two stages: 1) disentanglement learning, aiming to separate the semantic content and visual style of a given image; and 2) personalized modeling, which adapts LMMs to generate personalized images by integrating style patterns from user historically interacted images with the semantic information from the reference image.} 
\label{fig:dali}
\end{figure*}

To advance LMM-based personalized image generation, we propose explicitly disentangling style and semantics from both historical images and the reference image. This enables more effective personalization by better fusing the user’s preferred style preferences extracted from history and the reference semantics to guide image generation. Towards this, we introduce a novel method called DRC. As illustrated in Figure~\ref{fig:dali}, DRC consists of two key components:
\begin{itemize}[leftmargin=*]
    \item Disentanglement learning, which employs a self-supervised approach to disentangle the style and semantics of each image.
    \item Personalized modeling, which adapts LMMs to generate personalized images by integrating historical style patterns, the semantics of the reference image, and the user's textual instructions.
\end{itemize}

\subsection{Disentanglement Learning}
\label{sec:disen}
We introduce a dual-tower disentangler to separately extract style and semantics information, following a reconstruction paradigm closely aligned with LMM generation to enable effective disentanglement, as illustrated on the left side of Figure~\ref{fig:dali}. To ensure meaningful separation of the two targeted aspects, we first construct two types of contrastive samples for supervision: images that share the same style as the anchor (positive-style) and images that share the same semantics (positive-semantic). We use these samples to guide the disentanglement by enforcing that swapping the semantic representations between the anchor and the positive-semantic samples still enables successful image reconstruction—similarly for the style dimension.
In the following, we respectively detail the construction of contrastive samples, the dual-tower disentangler, the reconstruction process, and the training procedure.

\subsubsection{\textbf{Contrastive Samples}} 
\label{sec:contrastive}
To support meaningful disentanglement, for each anchor image $\bm{x}_a$, we construct: (1) a positive-style image $\bm{x}_s$ that shares the same style but differs in semantics, 
and (2) a positive-semantic image $\bm{x}_m$ that shares the same semantics but has a different style, 
using the following process:
\begin{itemize}[leftmargin=*]
    \item Positive-style image: Inspired by~\cite{huang2017arbitrary}, we employ VGG-Net~\cite{simonyan2014very} to extract style features from all images, compute style similarity using mean squared error, and select the most similar images from the candidate pool as the positive-style samples.
    
    \item Positive-semantic image: We use LLaVA~\cite{liu2023visual} to extract textual semantics of each anchor image, then employ SDXL~\cite{podell2024sdxl} to create images that retain the same semantics but differ in style.
\end{itemize}
Together with the anchor samples, these carefully constructed contrastive samples provide rich relational signals about the shared (also, the distinct) style and semantic patterns. 
Examples of the contrastive samples are presented in Appendix~\ref{sec:contrastive_example} for reference.

\subsubsection{\textbf{Dual-tower Disentangler}} 
We implement $\text{\textbf{StyleTower}}(\cdot)$ and $\text{\textbf{SemTower}}(\cdot)$ to extract style and semantic representations from input images, respectively. Both towers share the same architecture—comprising a cross-attention layer followed by a feed-forward layer—and operate in a similar manner. We use $\text{\textbf{StyleTower}}(\cdot)$ as an example to illustrate their functionality.

For an image triplet $<\bm{x}_a, \bm{x}_s, \bm{x}_m>$, we focus only on $\bm{x}_a$ and $\bm{x}_s$ when extracting style features. To extract the style from either image, the $\text{\textbf{StyleTower}}$ uses the target image as the query and the other as the key and value in the cross-attention layer. Let $\bm{F}_a^{\text{sty}}$ and $\bm{F}_s^{\text{sty}}$ denote the extracted style representation of $\bm{x}_a$ and $\bm{x}_s$, respectively. They can be formulated as follows:
\begin{equation}
\begin{split}
    \bm{F}_a^{\text{sty}} = \text{\textbf{StyleTower}}(q=\bm{E}_a,k=v=\bm{E}_s), \\
    \bm{F}_s^{\text{sty}} = \text{\textbf{StyleTower}}(q=\bm{E}_s,k=v=\bm{E}_a),
\end{split}
\end{equation}
where $q$, $k$, and $v$ denote the query, key, and value in the cross-attention. $\bm{E}_a$ and $\bm{E}_s$ represent the visual token embedding sequence of $\bm{x}_a$ and $\bm{x}_s$, obtained via the LMM’s visual tokenizer and embedding lookup. Since the query and key originate from images with similar style characteristics, the attention mechanism is expected to capture the shared style features.
Following a similar process, the $\text{\textbf{SemTower}}$ extracts the semantic information from $\bm{x}_a$ and $\bm{x}_m$, yielding semantic representations $\bm{F}_a^{\text{sem}}$ and $\bm{F}_m^{\text{sem}}$, respectively. 

\subsubsection{\textbf{Fusion for Reconstruction}}
With the disentangled representations, we reconstruct the original images through a recovery process. To align with the LMM generation pipeline, we first employ a fusion network to encode the combination of style and semantic features into compact latent instructions, which are then used to prompt the LMM for image generation.
Ideally, combining any style feature from $\{ \bm{F}_a^{\text{sty}}, \bm{F}_s^{\text{sty}} \}$ with any semantic feature from $\{ \bm{F}_a^{\text{sem}}, \bm{F}_m^{\text{sem}} \}$ could allow for the reconstruction of the original anchor image. Accordingly, all possible combinations are considered for image reconstruction.
Using the combination $(\bm{F}_a^{\text{sty}}, \bm{F}_m^{\text{sem}})$ as an example, the fusion process is formulated as follows:
\begin{equation}
    \bm{z}_{(a,m)} = \text{\textbf{FusionNetwork}}(sty=\bm{F}_a^{\text{sty}}, sem=\bm{F}_m^{\text{sem}}). \\
\end{equation}
Here, $\bm{z}_{(a,m)}$ denotes the latent instruction obtained for the combination of $(\bm{F}_a^{\text{sty}}, \bm{F}_m^{\text{sem}})$. The \text{\textbf{FusionNetwork}} consists of two cross-attention layers, where the two inputs alternately serve as query and key-value to capture intricate interdependencies, followed by residual connections and normalization to ensure stable and effective feature integration.
Similarly, we can obtain the latent instructions $\bm{z}_{(a,a)}$, $\bm{z}_{(s,a)}$, and $\bm{z}_{(s,m)}$ for $(\bm{F}_a^{\text{sty}}, \bm{F}_a^{\text{sem}})$, $(\bm{F}_s^{\text{sty}}, \bm{F}_a^{\text{sem}})$, and  $(\bm{F}_s^{\text{sty}}, \bm{F}_a^{\text{sem}})$, respectively.  
Each latent instruction is then separately inputted to the LMM for image generation.

\subsubsection{\textbf{Training}} 
\label{sec:disen_train}
During training, we enforce the model to reconstruct the original image using any combination from $\mathcal{Z} = \{\bm{z}{(a,a)}, \bm{z}{(a,m)}, \bm{z}{(s,a)}, \bm{z}{(s,m)}\}$. This encourages the StyleTower to extract the shared information—explicitly defined as style during the construction of contrastive samples—between the anchor and positive-style images, and the SemTower to capture the shared semantics between the anchor and positive-semantic images, thereby promoting meaningful disentanglement.

\vspace{3pt}
\noindent\textbf{$\bullet$ Training objective.}
To facilitate instruction tuning of the LMM, we design a hybrid prompt that integrates both latent and textual instructions, \textit{``Create an image that meets the specified requirements} $\bm{z}$'', where $\bm{z} \in \mathcal{Z}$ can be any the obtained latent instruction, optimizing the model to reconstruct anchor token sequence $\bm{E}_a$ via next-token prediction. The training objective is:
\begin{equation}
\begin{gathered}
    \mathcal{L}_{\text{recon}} = \mathbb{E}_{\bm{z}\sim P(\mathcal{Z})} \left[ \ell(\bm{z}) \right], \\
    \text{where}\ \ell(\bm{z}) = - \textstyle\sum_{j} \log \, P_\Theta \left( \bm{E}_{a,j} \,\middle|\ {\bm{z}}, \bm{E}_{a,<j} \right).
\end{gathered}
\end{equation}
Here, $P_\Theta \left( \bm{E}{a,j} ,\middle|\ {\bm{z}}, \bm{E}{a,<j} \right)$ denotes the predicted probability of the $j$-th visual token of the anchor image, conditioned on the latent instruction $\bm{z}$ and the preceding tokens, while $\ell(\bm{z})$ represents the reconstruction loss associated with using $\bm{z}$. The term $\bm{z}\sim P(\mathcal{Z})$ refers to sampling $\bm{z}$ from the set $\mathcal{Z}$ according to a fixed probability distribution.
Since different style-semantic combinations present varying levels of optimization difficulty, we adopt a \textbf{difficulty-aware importance sampling} strategy to prioritize more challenging combinations during training, rather than relying on uniform sampling. Specifically, we assign higher sampling probabilities to combinations with greater reconstruction losses, with the sampling probability for each $\bm{z}$ defined as follows:
\begin{equation}
    p(\bm{z}) = \frac{\ell(z)}{
    \sum_{z^{\prime} \in \mathcal{Z}} {\ell(z^{\prime)}}
    }. 
\end{equation}

\subsection{Personalized Modeling} 
Equipped with the well-trained disentanglement mechanism, DRC facilitates personalized image generation by constructing user-specific latent instructions from the historically interacted images and the given reference image to steer generation using the LMM, as shown in Figure~\ref{fig:dali} (b). The history images provide user-preferred style information, while the reference image supplies users' semantic intentions. 
Next, we describe how user-specific latent instructions are constructed for personalized image generation, followed by the design of the training and inference procedures.

\subsubsection{\textbf{User-specific Latent Instructions}} 
Given a user $u$'s historically interacted images $\mathcal{H}_u=\{\bm{x}_i\}_{i=1}^N$, and multimodal instructions $\mathcal{M}_u=\{{\bm{x}_{r}},txt\}$, where $\bm{x}_r$ and $txt$ denote the reference image and textual instruction, respectively, personalized image generation aims to capture user style preferences from the history images and integrate them with the semantic intent conveyed by the multimodal instructions, thereby guiding the generation of personalized target image $\bm{x}_{N+1}$. At its core, the task involves: (1) accurately extracting relevant style and semantic cues, and (2) effectively fusing them to produce latent instructions that guide the generation process.  We leverage the dual-tower disentangler and the fusion network developed during the disentanglement learning process to complete the tasks, as detailed below:

\vspace{3pt}
\noindent\textbf{$\bullet$ User-preferred style and reference semantics extraction.} 
We first leverage the trained \text{\textbf{StyleTower}} and \text{\textbf{SemTower}} to extract user-preferred style information from historical images and semantic information from the reference image, respectively. 
Originally, the two towers operate in a cross-attention manner using the anchor and contrastive images. Here, however, we adapt them to a self-attention setting—where the query, key, and value are all derived from the same source—to directly extract information from the historical or reference images, keeping similar to~\cite{hao2023learning}. Formally, the extracted user-preferred style information $\bm{F}^{\text{sty}}$ and  reference semantic information $\bm{F}^{\text{sem}}$ for the user are obtained as:
\begin{equation}
\begin{split}
\bm{F}^{\text{sty}}  = & \left[ \bm{F}_{1}^{\text{sty}}, \dots, \bm{F}_{i}^{\text{sty}}, \dots, \bm{F}_{N}^{\text{sty}} \right], \\
& \text{where} \, \bm{F}_{i}^{\text{sty}} =  \text{\textbf{StyleTower}}\left(q = \bm{E}_i, k/v = \bm{E}_i\right), \\
\bm{F}^{\text{sem}} = &  \text{\textbf{SemTower}}\left(q = \bm{E}_r, k/v = \bm{E}_{r}\right).
\end{split} 
\end{equation}
Here, $\bm{E}_{i}$ denotes the visual token embedding sequences corresponding to the $i$-th historical image $\bm{x}_{i}$, and $\bm{E}_{r}$ denotes the visual token embedding sequence for the reference image $\bm{x}_{r}$.

Given that user-preferred and disliked features (e.g., colors and patterns) often coexist in the history images, the extracted style information inevitably contain noise. Meanwhile, the semantic information govern the overall content and structure of the generated image, and dynamically adjusting their influence is crucial for balancing semantic alignment and personalized style preservation.
To achieve both robust noise filtering and flexible semantic control, we apply a learnable mask to control the final style and semantic information used. Additionally, we employ mean pooling to aggregate all historical style information. Formally, the final user-preferred style representation $\tilde{\bm{F}}^{\text{sty}}$ and reference semantic representation $\tilde{\bm{F}}^{\text{sem}}$ are obtained as follows:
\begin{equation}
    \tilde{\bm{F}}^{\text{sty}} = \frac{1}{N}\textstyle\sum_{i=1}^N(\bm{F}_{i}^{\text{sty}}\odot\bm{m}_{i}^{\text{sty}}), \quad \tilde{\bm{F}}^{\text{sem}} = \bm{F}^{\text{sem}} \odot \bm{m}^{\text{sem}},
\end{equation}
where $\bm{m}_{i}^{\text{sty}}$ and $\bm{m}^{\text{sem}}$ denote the learned masks for $\bm{F}_{i}^{\text{sty}}$ and $\bm{F}^{\text{sem}}$. Following~\cite{xu2024personalized}, the masks are generated using a transformer encoder-based mask generator that predicts binary masks. Formally,
\begin{equation}
    \bm{m}_{i}^{\text{sty}}, \dots, \bm{m}_{N}^{\text{sty}},  \bm{m}^{\text{sem}} = \text{\textbf{MaskGenerator}}(\bm{F}^{\text{sty}}, \bm{F}^{\text{sem}}, \alpha_s, \alpha_m),
\end{equation}
where $\alpha_s, \alpha_m\in [0,1]$ are the mask ratios for style and semantic information, respectively. By jointly processing style and semantic information, the mask generator enables cross-feature awareness, allowing the style mask to leverage semantic context for better noise filtering and the semantic mask to adaptively adjust the influence of the semantic information.

\vspace{3pt}
\noindent\textbf{$\bullet$ Obtaining Latent Instructions via Style and Semantic Fusion.} Finally, we fuse the obtained user-preferred style representation with the reference semantic representation to form the user-specific latent instruction $\tilde{\bm{z}}$ for the user $u$: 
\begin{equation}
    \tilde{\bm{z}} = \text{\textbf{FusionNetwork}}(\tilde{\bm{F}}^{\text{sty}}, \tilde{\bm{F}}^{\text{sem}}),
\end{equation}
which serves as the personalized condition to guide target image generation within the LMM.

% \vspace{-0.1cm}
\subsubsection{\textbf{Training}}
\label{sec:sem_aug}
Similar to the first-stage training, DRC integrates the user-specific latent instruction $\tilde{\bm{z}}$ with the textual instruction $txt$ to form the hybrid prompt, guiding the model to generate the personalized target image $\bm{E}_{N+1}$. To construct a high-quality supervised dataset for effective learning, we adopt semantic-preserving augmentation to diversify the reference images, then conduct supervised fine-tuning for personalized image generation.

\vspace{3pt}
\noindent\textbf{$\bullet$ Semantic-preserving augmentation.}  As depicted in Figure~\ref{fig:intro}, we construct the reference image set $\mathcal{R}=\{\bm{x}_{r}^k\}_{k}$ by applying semantic-preserving augmentation to the target image ($\bm{x}_{N+1}$) for more robust personalization. Specifically, we reuse the positive semantic image construction strategy outlined in Section~\ref{sec:contrastive} to synthesize multiple reference images that retain target semantics while exhibiting diverse styles.

\vspace{3pt}
\noindent\textbf{$\bullet$ Supervised Fine-Tuning (SFT).} Given the diverse interpretations of style and semantics across different users in personalized scenarios, DRC continuously optimizes all learnable parameters $\Theta$ to ensure effective alignment with diverse user style preferences and semantic intentions, with the objective formulated as:
\begin{equation}
    \resizebox{.9\hsize}{!}{$\mathcal{L}_{\text{sft}}=\mathbb{E}_{\substack{\bm{x}_r\sim \mathcal{R} \\ \alpha_m\sim\mathcal{U}(0,1)}}\left[-\textstyle\sum_j\log P_\Theta\left(\bm{E}_{N+1,j}\ \middle|\ txt, \tilde{\bm{z}}, \bm{E}_{N+1,<j}\right)\right],$}
\end{equation}
where $\bm{x}_{r} \sim \mathcal{R}$ denotes the random selection of one image from $\mathcal{R}$ as the reference image, $\alpha_m \sim \mathcal{U}(0,1)$ represents randomly sampling a mask ratio for semantic information during training, and $\tilde{\bm{z}}$ denotes the corresponding latent instruction obtained under $\alpha_m$. Additionally, $P_\Theta\left(\bm{E}_{N+1,j}\ \middle|\ \text{txt}, \tilde{\bm{z}}, \bm{E}_{N+1,<j}\right)$ denotes the predicted probability for the $j$-th token of the target image.

Notably, the reason for randomly sampling the mask ratio for the semantic part is to allow the model to flexibly control the influence of the reference semantic representations. The mask ratio for the style part is treated as a hyperparameter.

\subsubsection{\textbf{Inference}} To balance semantic alignment and personalized style preservation, users can adjust the semantic mask ratio $\alpha_m$ to control the semantics of the reference image to be incorporated into the generated images. During inference, given the history images $\mathcal{H}_u = \{\bm{x}_i\}_{i=1}^{N}$, the multimodal instruction $\mathcal{M}_u = \{\bm{x}_0, \mathit{txt}\}$, and the user-specified semantic mask ratio $\alpha_m$, DRC constructs the hybrid prompt consisting of $\{\tilde{\bm{z}, txt}\}$ to guide the generation of personalized visual token sequence $\bm{E}_{N+1}$, which are subsequently decoded to produce the final personalized image $\bm{x}_{N+1}$.

\section{Experiments}
\label{sec:experiment}

We evaluate the proposed DRC framework in both sticker and movie poster scenarios to answer the following research questions:

\begin{itemize}[leftmargin=*]
\item \textbf{RQ1:} How does DRC perform compared to other DM-based, LLM-based, and LMM-based methods across different scenarios?
\item \textbf{RQ2:} What is the contribution of each component in DRC (i.e., difficulty-aware importance sampling, FusionNetwork, and the disentangler architecture) to disentanglement learning?
\item \textbf{RQ3:} How do various mechanisms (i.e., supervised fine-tuning and semantic-preserving augmentation) in personalized modeling affect the effectiveness of  DRC?
\end{itemize}

\begin{table}[t]
\setlength{\abovecaptionskip}{0cm}
\setlength{\belowcaptionskip}{0.2cm}
\caption{Overview of the dataset statistics.}
\label{tab:dataset_info}
\begin{tabular}{l|cc}
\hline
\multicolumn{1}{l|}{}                       & \multicolumn{1}{l}{\textbf{Stickers}} & \multicolumn{1}{l}{\textbf{Movie posters}} \\ \hline
\textbf{\#Users}                           & 725                                   & 594                                        \\
\textbf{\#Items}                           & 14345                                 & 6961                                       \\
\textbf{\#Disentanglement} & 14984                                 & 50176                                      \\
\textbf{\#Personalized}             & 10719                                 & 31058                                      \\ \hline
\end{tabular}
\vspace{-0.4cm}
\end{table}

% \usepackage{multirow}
% \usepackage[normalem]{ulem}
\useunder{\uline}{\ul}{}
\begin{table*}[]
\setlength{\abovecaptionskip}{0.1cm}
\setlength{\belowcaptionskip}{0.1cm}
\caption{Quantitative evaluation results of DRC and other baselines. Models marked with ``*'' indicate the pre-trained models. The best-performing metrics are highlighted in bold, while the second-best metrics are \underline{underlined}.}
\label{tab:quantitative_evaluation}
\begin{tabular}{cl|cccccc|ccc|c}
\hline
\multicolumn{2}{l|}{\textbf{\#Sticker}}                                             & \multicolumn{6}{c|}{\textbf{Style Alignment}}                                                                     & \multicolumn{3}{c|}{\textbf{Semantic Alignment}}                    & \textbf{Fidelity}    \\ \hline
\multicolumn{2}{c|}{\textbf{Methods}}                                               & \textbf{CS $\uparrow$}         & \textbf{CIS$\uparrow$}        & \textbf{DIS$\uparrow$}        & \textbf{LPIPS$\downarrow$}      & \textbf{StyleSim$\downarrow$} & \textbf{CSD$\uparrow$}    & \textbf{CS$\uparrow$}         & \textbf{CIS$\uparrow$}        & \textbf{DIS$\uparrow$}         & \textbf{FID $\downarrow$}        \\ \hline
\textbf{DM-based}                                            & \textbf{TI}          & 18.67                & 40.90                & 36.58                & 0.7654               & 7.63 & 0.4883                & \textbf{32.91}       & {\ul 53.67}          & {\ul 48.50}           & 105.48               \\ \hline
\cellcolor[HTML]{FFFFFF}\textbf{LLM-based}                   & \textbf{PMG}         & 19.16                & 47.34                & 39.15                & 0.7383               & 13.94 & 0.4704                & 18.31                & 45.45                & 37.80                 & {\ul 84.91}          \\ \hline
\cellcolor[HTML]{FFFFFF}                                     & \textbf{LLaVA*}      & 17.88                & 47.26                & 42.59                & 0.7575               & 8.11 & 0.4399                 & 17.54                & 42.65                & 39.25                 & 93.23                \\
\cellcolor[HTML]{FFFFFF}                                     & \textbf{LLaVA}       & 18.72                & 37.44                & 33.19                & 0.7552               & 7.64 & 0.4991                 & {\ul 27.02}          & 49.15                & 43.88                 & 95.19                \\
\cellcolor[HTML]{FFFFFF}                                     & \textbf{LaVIT*}      & 18.77                & 53.63                & 50.96                & 0.6855               & 9.94 & 0.4732                 & 15.49                & 40.76                & 39.09                 & 107.53               \\
\cellcolor[HTML]{FFFFFF}                                     & \textbf{LaVIT}       & 16.39                & 40.56                & 40.84                & 0.7377               & 8.59 & 0.5919                 & 25.74                & \textbf{70.80}       & \textbf{69.93}        & \textbf{83.39}       \\
\cellcolor[HTML]{FFFFFF}                                     & \textbf{Pigeon}      & {\ul 22.03}          & {\ul 61.64}          & {\ul 57.26}          & {\ul 0.6800}         & {\ul 6.99} & {\ul 0.6531}           & 25.74                & 50.66                & 48.34                 & 93.60                \\
\rowcolor[HTML]{D9D9D9} 
\multirow{-6}{*}{\cellcolor[HTML]{FFFFFF}\textbf{LMM-based}} & \textbf{DRC}        & \textbf{23.19}       & \textbf{65.10}       & \textbf{60.00}       & \textbf{0.6770}      & \textbf{6.52} & \textbf{0.6820}        & 21.88                & 48.30                & 45.79                 & 92.31                \\ \hline
                                                             & \textbf{Recon} & 16.30                & 40.60                & 40.76                & 0.7370               & 8.84 & 0.4715                 & 25.84                & 71.09                & 70.14                 & 83.57                \\
\multirow{-2}{*}{\textbf{Reference}}                         & \textbf{Grd}         & 16.93                & 45.00                & 43.71                & 0.6443               & 8.10 & 0.5167                 & 28.95                & 100.00               & 100.00                & -                    \\ \hline
\end{tabular}\\

\begin{tabular}{cl|cccccc|ccc|c}
\hline
\multicolumn{2}{l|}{\textbf{\#Movie poster}}                                        & \multicolumn{6}{c|}{\textbf{Style Alignment}}                                                                     & \multicolumn{3}{c|}{\textbf{Semantic Alignment}}                    & \textbf{Fidelity}    \\ \hline
\multicolumn{2}{c|}{\textbf{Methods}}                                               & \textbf{CS $\uparrow$}         & \textbf{CIS $\uparrow$}        & \textbf{DIS $\uparrow$}        & \textbf{LPIPS $\downarrow$}      & \textbf{StyleSim $\downarrow$} & \textbf{CSD $\uparrow$}   & \textbf{CS $\uparrow$}         & \textbf{CIS $\uparrow$}        & \textbf{DIS $\uparrow$}         & \textbf{FID $\downarrow$}        \\ \hline
\textbf{DM-based}                                            & \textbf{TI}          & 12.41                & 28.29                & 19.18                & 0.7721               & 10.83 & 0.3042                & \textbf{33.84}       & 43.53                & 39.81                 & 79.77                \\ \hline
\cellcolor[HTML]{FFFFFF}\textbf{LLM-based}                   & \textbf{PMG}         & 13.61                & 25.11                & \textbf{22.73}       & 0.7692               & 16.69 & 0.3141                & 15.60                & 27.29                & 25.15                 & 77.25                \\ \hline
\cellcolor[HTML]{FFFFFF}                                     & \textbf{LLaVA*}      & 12.24                & 29.60                & {\ul 19.73}          & 0.7607               & 10.02 & 0.3152                & 14.55                & 31.76                & 21.99                 & 73.77                \\
\cellcolor[HTML]{FFFFFF}                                     & \textbf{LLaVA}       & 12.62                & 30.64                & 19.33                & 0.7690               & 10.67 & 0.3049                & {\ul 30.53}          & \textbf{48.50}       & 41.45                 & 54.55                \\
\cellcolor[HTML]{FFFFFF}                                     & \textbf{LaVIT*}      & 12.64                & 28.23                & 17.50                & 0.7546               & \textbf{8.87} & {\ul 0.3603}        & 19.39                & 36.93                & 37.71                 & 50.08                \\
\cellcolor[HTML]{FFFFFF}                                     & \textbf{LaVIT}       & 13.86                & 30.49                & 19.95                & 0.7548               & 10.97 & 0.3352                & 25.15                & 46.02                & \textbf{60.07}        & \textbf{33.53}       \\
\cellcolor[HTML]{FFFFFF}                                     & \textbf{Pigeon}      & {\ul 15.19}          & {\ul 37.42}          & 17.70                & \textbf{0.7496}      & 10.54 & 0.3473                & 27.30                & 47.79                & 41.35                 & {\ul 41.35}          \\
\rowcolor[HTML]{D9D9D9} 
\multirow{-6}{*}{\cellcolor[HTML]{FFFFFF}\textbf{LMM-based}} & \textbf{DRC}        & \textbf{15.24}       & \textbf{37.53}       & 19.26                & {\ul 0.7538}         & {\ul 9.57} & \textbf{0.3607}           & 25.71                & {\ul 48.26}          & {\ul 43.45}           & 42.20                \\ \hline
                                                             & \textbf{Recon} & 13.85                & 30.33                & 19.95                & 0.7548               & 9.82 & 0.3589                 & 25.29                & 46.08                & 60.52                 & 33.74                \\
\multirow{-2}{*}{\textbf{Reference}}                         & \textbf{Grd}         & 10.94                & 51.34                & 20.75                & 0.7502               & 9.89 & 0.3769                 & 31.81                & 100.00               & 100.00                & -                    \\ \hline
\end{tabular}
\vspace{-0.2cm}
\end{table*}

\subsection{Experimental Settings}
\subsubsection{\textbf{Datasets}}
We evaluate the proposed DRC on two publicly available datasets: 1) \textbf{SER30K}\footnote{\url{https://github.com/nku-shengzheliu/SER30K.}} is a sticker dataset containing a wide range of sticker themes and images. 2) \textbf{ML-Latest}\footnote{\url{https://grouplens.org/datasets/movielens.}} is a movie poster dataset comprising poster images and user rating information. 

For disentanglement learning, we follow Section~\ref{sec:contrastive} to construct style-positive and semantics-positive samples. For personalized modeling, we adopt the pipeline in~\cite{xu2024personalized}, where each sample includes user-interacted history images, a reference image, and a personalized target. The reference is generated via semantic-preserving augmentation on the target, as described in Section~\ref{sec:sem_aug}. All samples are split into training, validation, and testing sets with a ratio of 8:1:1. Dataset statistics are shown in Table~\ref{tab:dataset_info}.

\subsubsection{\textbf{Baselines}}
We compare DRC with several DM-based, LLM-based, and LMM-based methods: 1) \textbf{Textual Inversion} (TI) \cite{gal2023an} learns user preferences using a word embedding to guide image generation. 2) \textbf{PMG} \cite{shen2024pmg} converts images into keywords to capture user preferences and intentions for guiding image generation. 3) \textbf{LLaVA} \cite{liu2023visual} is a representative LMM for image understanding, which can be adapted for image generation with an external text-to-image generator. 4) \textbf{LaVIT} \cite{jin2024unified} is another LMM that quantizes images into discrete visual tokens for both image understanding and generation. 5) \textbf{Pigeon} \cite{xu2024personalized}, built upon LaVIT, introduces a dedicated mask generator to discard noises from history images for personalized image generation. For fair comparison, both DRC and Pigeon are evaluated under the supervised fine-tuning setting only.

Additionally, we include two additional results for comparison: 6) \textbf{Recon}, we tokenize the reference images using the visual tokenizer of LaVIT and directly reconstruct them without personalization. 7) \textbf{Grd}, the evaluation results of the reference images themselves.

\subsubsection{\textbf{Evaluation Metrics}} Following~\cite{xu2024personalized,shen2024pmg}, We adopt various evaluation metrics focusing on \textbf{style alignment} and \textbf{semantic alignment} between generated and history/reference images:

\begin{itemize}[leftmargin=*]
\item \textbf{Style Alignment.} To assess style preservation, we measure the visual similarity between generated images and user history images. Specifically, LPIPS~\cite{zhang2018unreasonable} is used to evaluate perceptual similarity, while StyleSim~\cite{huang2017arbitrary} and CSD~\cite{somepalli2024measuring} are used for style-level similarity. We further compute the CLIP Image Score (CIS) and DINO Image Score (DIS) by measuring the cosine similarity of visual features extracted by CLIP~\cite{radford2021learning} and DINO~\cite{oquab2023dinov2}, respectively. In addition, the CLIP Score (CS) is calculated between the generated images and textual descriptions of the history images.

\item \textbf{Semantic Alignment.} To assess the semantic consistency between the generated images and the reference images, we adopt the same CIS, DIS, and CS metrics while computing them between the generated and the reference images.

\item \textbf{Fidelity.} To evaluate the fidelity of the generated images, we adopt the Fréchet Inception Distance (FID) metric.
\end{itemize}

\begin{figure*}[t]
% \vspace{-0.2cm}
\setlength{\abovecaptionskip}{0.0cm}
\setlength{\belowcaptionskip}{-0.2cm}
\centering
\includegraphics[scale=0.56]{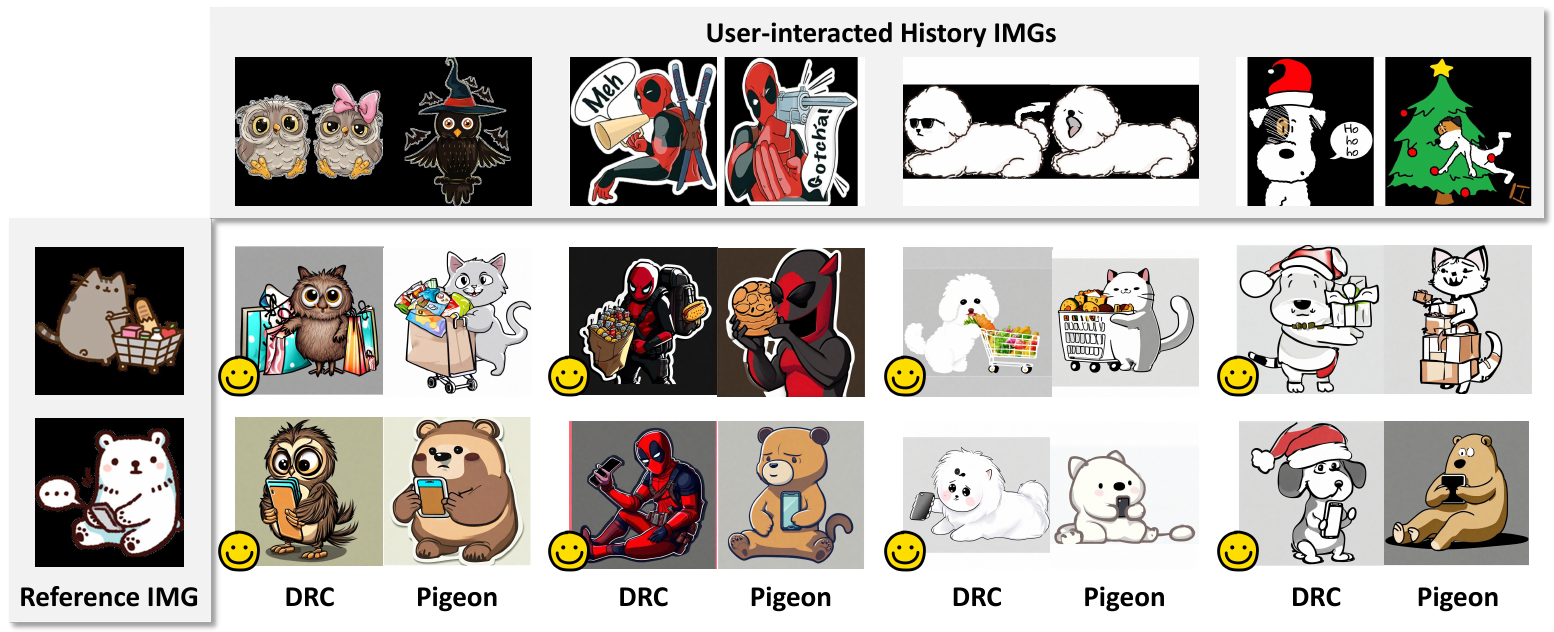}
\caption{Qualitative comparison between DRC and Pigeon.} 
\label{fig:disen_case}
\end{figure*}

\subsubsection{\textbf{Implementation Details}} All baselines are trained using their default settings. For DRC, we set the learning rates to $1e^{-4}$ and $1e^{-5}$ during the disentanglement learning and personalized modeling stages, respectively. Following \cite{xu2024personalized}, the style mask ratio $\alpha_s$ is fixed at $0.2$. During inference, we select the optimal semantic mask ratio $\alpha_m \in \{0.0, 0.1, \dots, 1.0\}$ for each image by averaging the style CIS and the semantic CS. All experiments are conducted on a single NVIDIA A100 GPU. Detailed computational costs of DRC are reported in Appendix~\ref{sec:cost}.

\subsection{Overall Performance (RQ1)}

% \noindent \textbf{Quantitative Evaluation.} 
We present the performance comparison between DRC and all baselines in Table~\ref{tab:quantitative_evaluation} with the observations as follows:

\begin{itemize}[leftmargin=*]
    \item DM-based TI outperforms other baselines in semantic alignment by effectively capturing the semantics of the reference image, while it struggles to extract user preferences from the history images, leading to unsatisfactory style alignment.
    \item PMG employs an LLM to capture user style preferences by converting images into textual descriptions. However, this image-to-text conversion often fails to effectively capture fine-grained visual details, leading to suboptimal performance in both personalization and semantic alignment.
    \item For the pre-trained LLaVA and LaVIT, it is challenging to effectively extract task-relevant information from complex history images and multimodal instructions, resulting in unsatisfactory performance. After fine-tuning, these models tend to focus on reconstructing the reference image while overlooking the style preferences in the user-interacted history images, which can be attributed to the lack of high-quality supervised datasets.
    \item DRC and Pigeon achieve the best overall performance, excelling in both style and semantic alignment. Notably, DRC achieves comparable, even superior performance to Pigeon in most metrics. The explicit disentanglement strategy allows the model to effectively separate and integrate style and semantic information, resulting in substantial performance gains. More importantly, as illustrated in Figure~\ref{fig:disen_case}, DRC can accurately extract and fuse user-preferred styles and reference semantics for personalized sticker generation, effectively mitigating the guidance collapse issue observed in Pigeon. For instance, in the first case, DRC successfully captures the owl character style from the user-interacted history images and combines it with the shopping semantics from the reference image to produce a shopping-themed owl sticker. In contrast, Pigeon leans toward reconstructing the reference image, neglecting the personalized style features present in the history. Additional qualitative examples are provided in Appendix~\ref{sec:cases}. Furthermore, we conduct a human evaluation to validate the effectiveness of DRC, with results presented in Appendix~\ref{sec:user_study}.
\end{itemize}

\subsection{In-depth Analysis}
To better analyze the effectiveness of our designs in disentanglement learning and personalized modeling, we conduct ablation studies on the sticker dataset for an in-depth analysis.
\subsubsection{\textbf{Effect of disentanglement learning (RQ2)}}

To investigate the impact of our designs in disentanglement learning, we conduct three ablation studies during training: 1) We remove the importance sampling strategy, referred to as “w/o Imp.” 2) We remove the FusionNetwork and instead concatenate the disentangled features directly, referred to as “w/o Fusion”. 3) We remove the attention-based disentangler and replace it with a simple MLP-based disentangler, which directly conducts token-level style-semantics disentanglement, referred to as “w/o Attn.”. The results are shown in Table~\ref{tab:disentanglement_analysis}, where we only present the representative metrics CIS for style alignment and CS for semantic alignment for brevity.

From the experimental results, we observe that: 1) Removing importance sampling leads to performance drops in both style and semantic alignment. This indicates that importance sampling facilitates more effective disentanglement training, enabling the model to extract relevant features more precisely. 2) After removing the FusionNetwork, we observe that the model struggles to fuse the style and semantic representations effectively and tends to rely more heavily on the semantic representations for image reconstruction. This indicates that the FusionNetwork plays a crucial role in integrating both task-relevant cues to produce personalized latent instructions. 3) The results under the “w/o Attn.” setting clearly demonstrate a substantial decline in style alignment. This is expected, as image style and semantics are globally entangled in the high-dimensional feature space, while the MLP-based token-level disentangler fails to capture these global dependencies within the image token sequence, ultimately resulting in ineffective disentanglement.

\begin{table}[t]
\setlength{\abovecaptionskip}{0.05cm}
\setlength{\belowcaptionskip}{0.2cm}
\caption{Effects of the designs in disentanglement learning. ``Imp.'' denotes importance sampling, ``Fusion'' refers to the fusion network, and``Attn.'' denotes the attention-based disentangler.}
\label{tab:disentanglement_analysis}
\begin{tabular}{l|cc|c}
\hline
                      & \multicolumn{2}{c|}{\textbf{Style Alignment}} & \textbf{Semantic Alignment} \\
\textbf{Method}       & \textbf{CIS} $\uparrow$         & \textbf{LPIPS} $\downarrow$       & \textbf{CS} $\uparrow$                \\ \hline
\textbf{DRC}         & 58.68        & 0.6840       & 19.48              \\
\textbf{- w/o Imp.}   & 58.07                 & 0.6856                  & 18.54                       \\
\textbf{- w/o Fusion} & 46.27                     & 0.6955                     & 23.43                           \\ 
\textbf{- w/o Attn.} & 44.11 & 0.7245 & 24.73 \\ \hline
\end{tabular}
\vspace{-0.3cm}
\end{table}

\subsubsection{\textbf{Effect of personalization modeling (RQ3)}}

\begin{figure}[t]
% \vspace{-0.2cm}
\setlength{\abovecaptionskip}{0.1cm}
\setlength{\belowcaptionskip}{-0.2cm}
\centering
\includegraphics[scale=0.36]{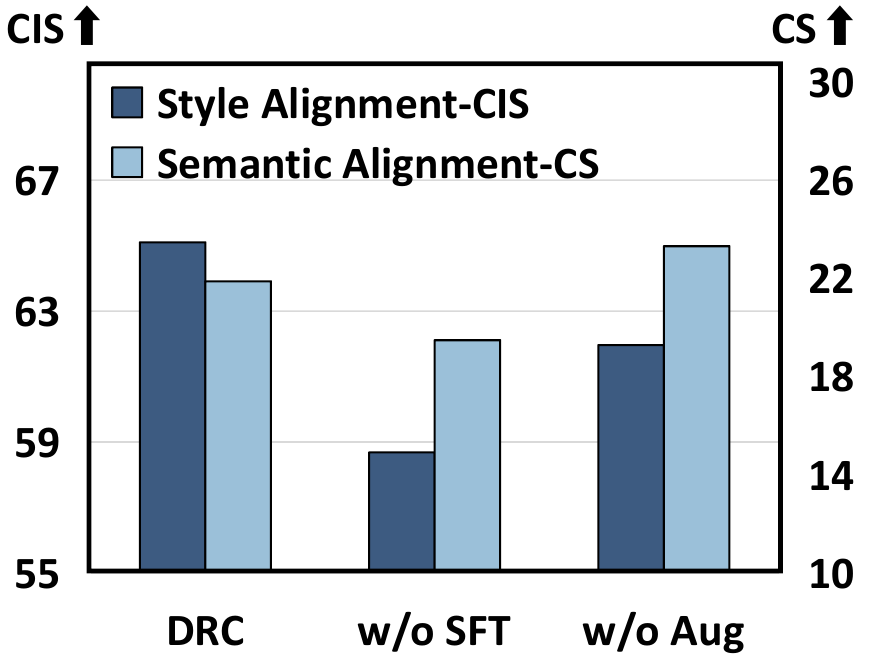}
\caption{In-depth analysis of SFT and the semantic-preserving augmentation during the second stage training, where ``w/o Aug'' means we directly utilize the target image as the reference.} 
\label{fig:personalized_analysis}
\end{figure}

To validate the mechanisms designed for the personalized modeling stage, we first evaluate the effectiveness of SFT by comparing model performance before and after this training step. As shown in Figure~\ref{fig:personalized_analysis}, both style alignment (CIS) and semantic alignment (CS) improve significantly after SFT, indicating that it effectively enhances the model’s ability to extract reference semantics, capture user-specific preferences, and integrate them. 

Then, we assess the effectiveness of semantic-preserving augmentation in personalized modeling, as shown in Figure~\ref{fig:personalized_analysis}.
Following Pigeon~\cite{xu2024personalized}, we conduct a variant where only the target image is used as the reference, referred to as ``w/o Aug''. 
While this mechanism results in marginally reduced semantic alignment, it substantially enhances the model's style alignment for personalized image generation.
This highlights the synthesized dataset's crucial role in enabling our DRC model to effectively balance semantic alignment and style alignment for personalized modeling.

\section{Related Work}
\label{sec:related_work}

\noindent\textbf{$\bullet$ Disentanglement Learning for Image Generation.} The inherently entangled nature of visual features often hinders controllable image generation, motivating extensive research into disentangled representation learning to enhance image synthesis across various scenarios, such as style transfer~\cite{qi2024deadiff}, pose-guided human generation~\cite{ma2018disentangled,xu2024disentangled,wang2024disco}, and face generation~\cite{deng2020disentangled,wei2024masterweaver}.
For example, DEADiff~\cite{qi2024deadiff} leverages Q-Former~\cite{li2023blip} to disentangle style and content representations by conditioning the encoder on ``style'' and ``content'' prompts, respectively, and injects these representations into mutually exclusive subsets of cross-attention layers within DMs, thereby enabling effective style transfer. ADI~\cite{huang2024learning} learns disentangled action identifiers by generating gradient masks to block the influence of action-agnostic features at the gradient level, guiding customized action generation. Inspired by these advances, we propose DRC, which disentangles image style and semantics and fuses task-relevant representations to steer more effective personalized image generation.

\vspace{3pt}
\noindent\textbf{$\bullet$ Personalized Generation.} With the rapid development of large generative models, such as DMs~\cite{podell2024sdxl,esser2024scaling}, LLMs~\cite{grattafiori2024llama,guo2025deepseek,yang2024qwen2}, and LMMs~\cite{zhou2025transfusion,chen2025janus,xie2025showo}, personalized generation has attracted growing attention across diverse modalities, including text~\cite{salemi-etal-2024-lamp,qiu2025measuring}, image~\cite{kang2025flux,li2025visualcloze,xiao2025omnigen,wu2025core}, 3D~\cite{raj2023dreambooth3d,liu2024make}, video~\cite{wang2024disco,wei2024dreamvideo,wu2024videomaker}, audio~\cite{li2024style,plitsis2024investigating}, and cross-modal generation~\cite{pham2024personalized,nguyen2024yollava}. By synthesizing content that aligns with specific user preferences and requirements, personalized generation paves the way for more engaging and user-centric experiences across various application scenarios~\cite{xu2025personalized}. 

In particular, personalized image generation stands out as a particularly active research avenue due to the rich expressiveness and wide applicability of visual content. Existing studies mainly fall into three lines: 1) DM-based methods such as TI~\cite{gal2023an} and DreamBooth~\cite{ruiz2023dreambooth}, mainly focus on subject-driven image generation guided by users' explicit multimodal instructions, while largely neglecting the modeling of user preferences derived from interaction history. In contrast, methods like DiFashion~\cite{xu2024diffusion} and I-AM-G~\cite{wang2024g} attempt to incorporate user-interacted history images to guide generation within DMs. However, these approaches face challenges when dealing with complex multimodal instructions. 2) LLM-based methods, such as PMG~\cite{shen2024pmg}, aim to capture user visual preferences by converting history images into text descriptions or directly using associated prompts, which inevitably results in the loss of visual details essential for accurate personalization. 3) LMM-based Pigeon~\cite{xu2024personalized} directly transforms history images and multimodal instructions into tokens to guide generation within LMMs, while the entanglement of visual features hinders the effectiveness of Pigeon in extracting useful information for generation. In this work, we introduce a style-semantics disentanglement mechanism to explicitly separate user style preferences and semantic intentions to guide LMM-based personalized image generation.

\vspace{3pt}
\noindent\textbf{$\bullet$ Large Multimodal Models.} Recently, an increasing number of studies have explored LMMs for visual understanding and generation. Some approaches, such as LLaVA~\cite{liu2023visual} and Mini-GPT4~\cite{zhu2024minigpt}, leverage pre-trained vision encoders to extract visual features, which are then projected into LLMs via a dedicated mapping layer for image understanding.
To enable both visual understanding and generation, models like LaVIT~\cite{jin2024unified} and SEED-X~\cite{ge2024seed} introduce dedicated visual tokenizers that convert images into discrete visual tokens through a learnable codebook, supporting autoregressive generation of both textual and visual tokens. In contrast, DreamLLM~\cite{dong2024dreamllm} enhances LLMs by introducing a special token to indicate image generation positions, along with a series of queries that encode historical semantics to guide the image generation process within a diffusion model-based decoder.
More recently, advanced LMMs with native image generation capabilities, such as Chameleon~\cite{team2024chameleon}, Show-o~\cite{xie2025showo}, Transfusion~\cite{zhou2025transfusion}, GPT-4o~\cite{openai2024gpt4o}, and BAGEL~\cite{deng2025emerging}, have demonstrated impressive performance in multimodal understanding and generation, establishing a solid foundation for future advancements in personalized image generation. Notably, our proposed DRC adopts LaVIT as its backbone, while its design remains compatible with other LMM architectures, which we leave as an exciting direction for future exploration.

\section{Conclusion and Future Work}
\label{sec:conclusion}

In this work, we identify \textit{Guidance Collapse} as a critical issue in personalized image generation, where entangled visual features prevent the accurate preservation of user-preferred styles and semantic intentions. To address this, we propose DRC, a novel framework that formulates user-specific latent instructions by explicitly disentangling style and semantics, thereby effectively guiding image generation within LMMs. Extensive experiments on personalized sticker and movie poster scenarios demonstrate that DRC consistently outperforms strong baselines across most metrics while effectively mitigating the guidance collapse issue. These findings underscore the importance of disentangled representation learning in enhancing controllability and effectiveness for personalized image generation.

Moving forward, several promising directions can be explored: 1) Extending DRC to integrate with a broader range of more LMMs~\cite{xie2025showo,zhou2025transfusion} to evaluate its generalizability and compatibility across diverse model architectures; 2) Exploring the applicability of DRC in more personalized generation scenarios, such as product image generation and digital avatar creation, to further validate its versatility and robustness in real-world applications; 3) Incorporating mechanisms to model evolving user preferences over time, enabling DRC to dynamically adapt to users' shifting style preferences and semantic intentions.

\vspace{5pt}

% \clearpage

% \section*{Acknowledgments}
% This work is supported by the National Natural Science Foundation of China (62272437).

{
\tiny
\bibliographystyle{ACM-Reference-Format}
\balance
\bibliography{bibfile}


\begin{thebibliography}{56}

%%% the Web address.

\ifx \showCODEN    \undefined \def \showCODEN     #1{\unskip}     \fi
\ifx \showISBNx    \undefined \def \showISBNx     #1{\unskip}     \fi
\ifx \showISBNxiii \undefined \def \showISBNxiii  #1{\unskip}     \fi
\ifx \showISSN     \undefined \def \showISSN      #1{\unskip}     \fi
\ifx \showLCCN     \undefined \def \showLCCN      #1{\unskip}     \fi
\ifx \shownote     \undefined \def \shownote      #1{#1}          \fi
\ifx \showarticletitle \undefined \def \showarticletitle #1{#1}   \fi
\ifx \showURL      \undefined \def \showURL       {\relax}        \fi
% invisible to TeX
\providecommand\bibfield[2]{#2}
\providecommand\bibinfo[2]{#2}
\providecommand\natexlab[1]{#1}
\providecommand\showeprint[2][]{arXiv:#2}

\bibitem[Chen et~al\mbox{.}(2025)]%
        {chen2025janus}
\bibfield{author}{\bibinfo{person}{Xiaokang Chen}, \bibinfo{person}{Zhiyu Wu}, \bibinfo{person}{Xingchao Liu}, \bibinfo{person}{Zizheng Pan}, \bibinfo{person}{Wen Liu}, \bibinfo{person}{Zhenda Xie}, \bibinfo{person}{Xingkai Yu}, {and} \bibinfo{person}{Chong Ruan}.} \bibinfo{year}{2025}\natexlab{}.
\newblock \showarticletitle{Janus-pro: Unified multimodal understanding and generation with data and model scaling}.
\newblock \bibinfo{journal}{\emph{arXiv:2501.17811}} (\bibinfo{year}{2025}).
\newblock


\bibitem[Chen et~al\mbox{.}(2024)]%
        {chen2024tailored}
\bibfield{author}{\bibinfo{person}{Zijie Chen}, \bibinfo{person}{Lichao Zhang}, \bibinfo{person}{Fangsheng Weng}, \bibinfo{person}{Lili Pan}, {and} \bibinfo{person}{Zhenzhong Lan}.} \bibinfo{year}{2024}\natexlab{}.
\newblock \showarticletitle{Tailored visions: Enhancing text-to-image generation with personalized prompt rewriting}. In \bibinfo{booktitle}{\emph{CVPR}}. \bibinfo{pages}{7727--7736}.
\newblock


\bibitem[Deng et~al\mbox{.}(2025)]%
        {deng2025emerging}
\bibfield{author}{\bibinfo{person}{Chaorui Deng}, \bibinfo{person}{Deyao Zhu}, \bibinfo{person}{Kunchang Li}, \bibinfo{person}{Chenhui Gou}, \bibinfo{person}{Feng Li}, \bibinfo{person}{Zeyu Wang}, \bibinfo{person}{Shu Zhong}, \bibinfo{person}{Weihao Yu}, \bibinfo{person}{Xiaonan Nie}, \bibinfo{person}{Ziang Song}, {et~al\mbox{.}}} \bibinfo{year}{2025}\natexlab{}.
\newblock \showarticletitle{Emerging properties in unified multimodal pretraining}.
\newblock \bibinfo{journal}{\emph{arXiv:2505.14683}} (\bibinfo{year}{2025}).
\newblock


\bibitem[Deng et~al\mbox{.}(2020)]%
        {deng2020disentangled}
\bibfield{author}{\bibinfo{person}{Yu Deng}, \bibinfo{person}{Jiaolong Yang}, \bibinfo{person}{Dong Chen}, \bibinfo{person}{Fang Wen}, {and} \bibinfo{person}{Xin Tong}.} \bibinfo{year}{2020}\natexlab{}.
\newblock \showarticletitle{Disentangled and controllable face image generation via 3d imitative-contrastive learning}. In \bibinfo{booktitle}{\emph{CVPR}}. \bibinfo{pages}{5154--5163}.
\newblock


\bibitem[Dong et~al\mbox{.}(2024)]%
        {dong2024dreamllm}
\bibfield{author}{\bibinfo{person}{Runpei Dong}, \bibinfo{person}{Chunrui Han}, \bibinfo{person}{Yuang Peng}, \bibinfo{person}{Zekun Qi}, \bibinfo{person}{Zheng Ge}, \bibinfo{person}{Jinrong Yang}, \bibinfo{person}{Liang Zhao}, \bibinfo{person}{Jianjian Sun}, \bibinfo{person}{Hongyu Zhou}, \bibinfo{person}{Haoran Wei}, \bibinfo{person}{Xiangwen Kong}, \bibinfo{person}{Xiangyu Zhang}, \bibinfo{person}{Kaisheng Ma}, {and} \bibinfo{person}{Li Yi}.} \bibinfo{year}{2024}\natexlab{}.
\newblock \showarticletitle{Dream{LLM}: Synergistic Multimodal Comprehension and Creation}. In \bibinfo{booktitle}{\emph{ICLR}}.
\newblock


\bibitem[Esser et~al\mbox{.}(2024)]%
        {esser2024scaling}
\bibfield{author}{\bibinfo{person}{Patrick Esser}, \bibinfo{person}{Sumith Kulal}, \bibinfo{person}{Andreas Blattmann}, \bibinfo{person}{Rahim Entezari}, \bibinfo{person}{Jonas M{\"u}ller}, \bibinfo{person}{Harry Saini}, \bibinfo{person}{Yam Levi}, \bibinfo{person}{Dominik Lorenz}, \bibinfo{person}{Axel Sauer}, \bibinfo{person}{Frederic Boesel}, {et~al\mbox{.}}} \bibinfo{year}{2024}\natexlab{}.
\newblock \showarticletitle{Scaling rectified flow transformers for high-resolution image synthesis}. In \bibinfo{booktitle}{\emph{ICML}}.
\newblock


\bibitem[Gal et~al\mbox{.}(2023)]%
        {gal2023an}
\bibfield{author}{\bibinfo{person}{Rinon Gal}, \bibinfo{person}{Yuval Alaluf}, \bibinfo{person}{Yuval Atzmon}, \bibinfo{person}{Or Patashnik}, \bibinfo{person}{Amit~Haim Bermano}, \bibinfo{person}{Gal Chechik}, {and} \bibinfo{person}{Daniel Cohen-or}.} \bibinfo{year}{2023}\natexlab{}.
\newblock \showarticletitle{An Image is Worth One Word: Personalizing Text-to-Image Generation using Textual Inversion}. In \bibinfo{booktitle}{\emph{ICLR}}.
\newblock


\bibitem[Ge et~al\mbox{.}(2024)]%
        {ge2024seed}
\bibfield{author}{\bibinfo{person}{Yuying Ge}, \bibinfo{person}{Sijie Zhao}, \bibinfo{person}{Jinguo Zhu}, \bibinfo{person}{Yixiao Ge}, \bibinfo{person}{Kun Yi}, \bibinfo{person}{Lin Song}, \bibinfo{person}{Chen Li}, \bibinfo{person}{Xiaohan Ding}, {and} \bibinfo{person}{Ying Shan}.} \bibinfo{year}{2024}\natexlab{}.
\newblock \showarticletitle{Seed-x: Multimodal models with unified multi-granularity comprehension and generation}.
\newblock \bibinfo{journal}{\emph{arXiv:2404.14396}} (\bibinfo{year}{2024}).
\newblock


\bibitem[Grattafiori et~al\mbox{.}(2024)]%
        {grattafiori2024llama}
\bibfield{author}{\bibinfo{person}{Aaron Grattafiori}, \bibinfo{person}{Abhimanyu Dubey}, \bibinfo{person}{Abhinav Jauhri}, \bibinfo{person}{Abhinav Pandey}, \bibinfo{person}{Abhishek Kadian}, \bibinfo{person}{Ahmad Al-Dahle}, \bibinfo{person}{Aiesha Letman}, \bibinfo{person}{Akhil Mathur}, \bibinfo{person}{Alan Schelten}, \bibinfo{person}{Alex Vaughan}, {et~al\mbox{.}}} \bibinfo{year}{2024}\natexlab{}.
\newblock \showarticletitle{The llama 3 herd of models}.
\newblock \bibinfo{journal}{\emph{arXiv:2407.21783}} (\bibinfo{year}{2024}).
\newblock


\bibitem[Guo et~al\mbox{.}(2025)]%
        {guo2025deepseek}
\bibfield{author}{\bibinfo{person}{Daya Guo}, \bibinfo{person}{Dejian Yang}, \bibinfo{person}{Haowei Zhang}, \bibinfo{person}{Junxiao Song}, \bibinfo{person}{Ruoyu Zhang}, \bibinfo{person}{Runxin Xu}, \bibinfo{person}{Qihao Zhu}, \bibinfo{person}{Shirong Ma}, \bibinfo{person}{Peiyi Wang}, \bibinfo{person}{Xiao Bi}, {et~al\mbox{.}}} \bibinfo{year}{2025}\natexlab{}.
\newblock \showarticletitle{Deepseek-r1: Incentivizing reasoning capability in llms via reinforcement learning}.
\newblock \bibinfo{journal}{\emph{arXiv:2501.12948}} (\bibinfo{year}{2025}).
\newblock


\bibitem[Hao et~al\mbox{.}(2023)]%
        {hao2023learning}
\bibfield{author}{\bibinfo{person}{Shaozhe Hao}, \bibinfo{person}{Kai Han}, {and} \bibinfo{person}{Kwan-Yee~K Wong}.} \bibinfo{year}{2023}\natexlab{}.
\newblock \showarticletitle{Learning attention as disentangler for compositional zero-shot learning}. In \bibinfo{booktitle}{\emph{CVPR}}. \bibinfo{pages}{15315--15324}.
\newblock


\bibitem[Huang et~al\mbox{.}(2024)]%
        {huang2024learning}
\bibfield{author}{\bibinfo{person}{Siteng Huang}, \bibinfo{person}{Biao Gong}, \bibinfo{person}{Yutong Feng}, \bibinfo{person}{Xi Chen}, \bibinfo{person}{Yuqian Fu}, \bibinfo{person}{Yu Liu}, {and} \bibinfo{person}{Donglin Wang}.} \bibinfo{year}{2024}\natexlab{}.
\newblock \showarticletitle{Learning disentangled identifiers for action-customized text-to-image generation}. In \bibinfo{booktitle}{\emph{CVPR}}. \bibinfo{pages}{7797--7806}.
\newblock


\bibitem[Huang and Belongie(2017)]%
        {huang2017arbitrary}
\bibfield{author}{\bibinfo{person}{Xun Huang} {and} \bibinfo{person}{Serge Belongie}.} \bibinfo{year}{2017}\natexlab{}.
\newblock \showarticletitle{Arbitrary style transfer in real-time with adaptive instance normalization}. In \bibinfo{booktitle}{\emph{ICCV}}. \bibinfo{pages}{1501--1510}.
\newblock


\bibitem[Jin et~al\mbox{.}(2024)]%
        {jin2024unified}
\bibfield{author}{\bibinfo{person}{Yang Jin}, \bibinfo{person}{Kun Xu}, \bibinfo{person}{Kun Xu}, \bibinfo{person}{Liwei Chen}, \bibinfo{person}{Chao Liao}, \bibinfo{person}{Jianchao Tan}, \bibinfo{person}{Yadong Mu}, {et~al\mbox{.}}} \bibinfo{year}{2024}\natexlab{}.
\newblock \showarticletitle{Unified Language-Vision Pretraining in LLM with Dynamic Discrete Visual Tokenization}. In \bibinfo{booktitle}{\emph{ICLR}}.
\newblock


\bibitem[Kang et~al\mbox{.}(2025)]%
        {kang2025flux}
\bibfield{author}{\bibinfo{person}{Hao Kang}, \bibinfo{person}{Stathi Fotiadis}, \bibinfo{person}{Liming Jiang}, \bibinfo{person}{Qing Yan}, \bibinfo{person}{Yumin Jia}, \bibinfo{person}{Zichuan Liu}, \bibinfo{person}{Min~Jin Chong}, {and} \bibinfo{person}{Xin Lu}.} \bibinfo{year}{2025}\natexlab{}.
\newblock \showarticletitle{Flux Already Knows--Activating Subject-Driven Image Generation without Training}.
\newblock \bibinfo{journal}{\emph{arXiv:2504.11478}} (\bibinfo{year}{2025}).
\newblock


\bibitem[Li et~al\mbox{.}(2023)]%
        {li2023blip}
\bibfield{author}{\bibinfo{person}{Junnan Li}, \bibinfo{person}{Dongxu Li}, \bibinfo{person}{Silvio Savarese}, {and} \bibinfo{person}{Steven Hoi}.} \bibinfo{year}{2023}\natexlab{}.
\newblock \showarticletitle{Blip-2: Bootstrapping language-image pre-training with frozen image encoders and large language models}. In \bibinfo{booktitle}{\emph{ICML}}. PMLR, \bibinfo{pages}{19730--19742}.
\newblock


\bibitem[Li et~al\mbox{.}(2024)]%
        {li2024style}
\bibfield{author}{\bibinfo{person}{Yinghao~Aaron Li}, \bibinfo{person}{Xilin Jiang}, \bibinfo{person}{Jordan Darefsky}, \bibinfo{person}{Ge Zhu}, {and} \bibinfo{person}{Nima Mesgarani}.} \bibinfo{year}{2024}\natexlab{}.
\newblock \showarticletitle{Style-talker: Finetuning audio language model and style-based text-to-speech model for fast spoken dialogue generation}.
\newblock \bibinfo{journal}{\emph{arXiv:2408.11849}} (\bibinfo{year}{2024}).
\newblock


\bibitem[Li et~al\mbox{.}(2025)]%
        {li2025visualcloze}
\bibfield{author}{\bibinfo{person}{Zhong-Yu Li}, \bibinfo{person}{Ruoyi Du}, \bibinfo{person}{Juncheng Yan}, \bibinfo{person}{Le Zhuo}, \bibinfo{person}{Zhen Li}, \bibinfo{person}{Peng Gao}, \bibinfo{person}{Zhanyu Ma}, {and} \bibinfo{person}{Ming-Ming Cheng}.} \bibinfo{year}{2025}\natexlab{}.
\newblock \showarticletitle{VisualCloze: A Universal Image Generation Framework via Visual In-Context Learning}.
\newblock \bibinfo{journal}{\emph{arXiv:2504.07960}} (\bibinfo{year}{2025}).
\newblock


\bibitem[Liu et~al\mbox{.}(2024)]%
        {liu2024make}
\bibfield{author}{\bibinfo{person}{Fangfu Liu}, \bibinfo{person}{Hanyang Wang}, \bibinfo{person}{Weiliang Chen}, \bibinfo{person}{Haowen Sun}, {and} \bibinfo{person}{Yueqi Duan}.} \bibinfo{year}{2024}\natexlab{}.
\newblock \showarticletitle{Make-your-3d: Fast and consistent subject-driven 3d content generation}. In \bibinfo{booktitle}{\emph{ECCV}}. Springer, \bibinfo{pages}{389--406}.
\newblock


\bibitem[Liu et~al\mbox{.}(2023)]%
        {liu2023visual}
\bibfield{author}{\bibinfo{person}{Haotian Liu}, \bibinfo{person}{Chunyuan Li}, \bibinfo{person}{Qingyang Wu}, {and} \bibinfo{person}{Yong~Jae Lee}.} \bibinfo{year}{2023}\natexlab{}.
\newblock \showarticletitle{Visual instruction tuning}.
\newblock \bibinfo{journal}{\emph{NeurIPS}}  \bibinfo{volume}{36} (\bibinfo{year}{2023}), \bibinfo{pages}{34892--34916}.
\newblock


\bibitem[Ma et~al\mbox{.}(2018)]%
        {ma2018disentangled}
\bibfield{author}{\bibinfo{person}{Liqian Ma}, \bibinfo{person}{Qianru Sun}, \bibinfo{person}{Stamatios Georgoulis}, \bibinfo{person}{Luc Van~Gool}, \bibinfo{person}{Bernt Schiele}, {and} \bibinfo{person}{Mario Fritz}.} \bibinfo{year}{2018}\natexlab{}.
\newblock \showarticletitle{Disentangled person image generation}. In \bibinfo{booktitle}{\emph{CVPR}}. \bibinfo{pages}{99--108}.
\newblock


\bibitem[Nguyen et~al\mbox{.}(2024)]%
        {nguyen2024yollava}
\bibfield{author}{\bibinfo{person}{Thao Nguyen}, \bibinfo{person}{Haotian Liu}, \bibinfo{person}{Yuheng Li}, \bibinfo{person}{Mu Cai}, \bibinfo{person}{Utkarsh Ojha}, {and} \bibinfo{person}{Yong~Jae Lee}.} \bibinfo{year}{2024}\natexlab{}.
\newblock \showarticletitle{Yo'{LL}a{VA}: Your Personalized Language and Vision Assistant}. In \bibinfo{booktitle}{\emph{NeurIPS}}.
\newblock


\bibitem[{OpenAI}(2024)]%
        {openai2024gpt4o}
\bibfield{author}{\bibinfo{person}{{OpenAI}}.} \bibinfo{year}{2024}\natexlab{}.
\newblock \bibinfo{title}{{GPT-4o Image Generation System Card Addendum}}.
\newblock \bibinfo{howpublished}{\url{https://openai.com/index/gpt-4o-image-generation-system-card-addendum/}}.
\newblock


\bibitem[Oquab et~al\mbox{.}(2023)]%
        {oquab2023dinov2}
\bibfield{author}{\bibinfo{person}{Maxime Oquab}, \bibinfo{person}{Timoth{\'e}e Darcet}, \bibinfo{person}{Th{\'e}o Moutakanni}, \bibinfo{person}{Huy Vo}, \bibinfo{person}{Marc Szafraniec}, \bibinfo{person}{Vasil Khalidov}, \bibinfo{person}{Pierre Fernandez}, \bibinfo{person}{Daniel Haziza}, \bibinfo{person}{Francisco Massa}, \bibinfo{person}{Alaaeldin El-Nouby}, {et~al\mbox{.}}} \bibinfo{year}{2023}\natexlab{}.
\newblock \showarticletitle{Dinov2: Learning robust visual features without supervision}.
\newblock \bibinfo{journal}{\emph{arXiv:2304.07193}} (\bibinfo{year}{2023}).
\newblock


\bibitem[Pham et~al\mbox{.}(2024)]%
        {pham2024personalized}
\bibfield{author}{\bibinfo{person}{Chau Pham}, \bibinfo{person}{Hoang Phan}, \bibinfo{person}{David Doermann}, {and} \bibinfo{person}{Yunjie Tian}.} \bibinfo{year}{2024}\natexlab{}.
\newblock \showarticletitle{Personalized Large Vision-Language Models}.
\newblock \bibinfo{journal}{\emph{arXiv:2412.17610}} (\bibinfo{year}{2024}).
\newblock


\bibitem[Plitsis et~al\mbox{.}(2024)]%
        {plitsis2024investigating}
\bibfield{author}{\bibinfo{person}{Manos Plitsis}, \bibinfo{person}{Theodoros Kouzelis}, \bibinfo{person}{Georgios Paraskevopoulos}, \bibinfo{person}{Vassilis Katsouros}, {and} \bibinfo{person}{Yannis Panagakis}.} \bibinfo{year}{2024}\natexlab{}.
\newblock \showarticletitle{Investigating personalization methods in text to music generation}. In \bibinfo{booktitle}{\emph{ICASSP}}. IEEE, \bibinfo{pages}{1081--1085}.
\newblock


\bibitem[Podell et~al\mbox{.}(2024)]%
        {podell2024sdxl}
\bibfield{author}{\bibinfo{person}{Dustin Podell}, \bibinfo{person}{Zion English}, \bibinfo{person}{Kyle Lacey}, \bibinfo{person}{Andreas Blattmann}, \bibinfo{person}{Tim Dockhorn}, \bibinfo{person}{Jonas M{\"u}ller}, \bibinfo{person}{Joe Penna}, {and} \bibinfo{person}{Robin Rombach}.} \bibinfo{year}{2024}\natexlab{}.
\newblock \showarticletitle{{SDXL}: Improving Latent Diffusion Models for High-Resolution Image Synthesis}. In \bibinfo{booktitle}{\emph{ICLR}}.
\newblock


\bibitem[Qi et~al\mbox{.}(2024)]%
        {qi2024deadiff}
\bibfield{author}{\bibinfo{person}{Tianhao Qi}, \bibinfo{person}{Shancheng Fang}, \bibinfo{person}{Yanze Wu}, \bibinfo{person}{Hongtao Xie}, \bibinfo{person}{Jiawei Liu}, \bibinfo{person}{Lang Chen}, \bibinfo{person}{Qian He}, {and} \bibinfo{person}{Yongdong Zhang}.} \bibinfo{year}{2024}\natexlab{}.
\newblock \showarticletitle{Deadiff: An efficient stylization diffusion model with disentangled representations}. In \bibinfo{booktitle}{\emph{CVPR}}. \bibinfo{pages}{8693--8702}.
\newblock


\bibitem[Qiu et~al\mbox{.}(2025)]%
        {qiu2025measuring}
\bibfield{author}{\bibinfo{person}{Yilun Qiu}, \bibinfo{person}{Xiaoyan Zhao}, \bibinfo{person}{Yang Zhang}, \bibinfo{person}{Yimeng Bai}, \bibinfo{person}{Wenjie Wang}, \bibinfo{person}{Hong Cheng}, \bibinfo{person}{Fuli Feng}, {and} \bibinfo{person}{Tat-Seng Chua}.} \bibinfo{year}{2025}\natexlab{}.
\newblock \showarticletitle{Measuring What Makes You Unique: Difference-Aware User Modeling for Enhancing LLM Personalization}.
\newblock \bibinfo{journal}{\emph{arXiv:2503.02450}} (\bibinfo{year}{2025}).
\newblock


\bibitem[Radford et~al\mbox{.}(2021)]%
        {radford2021learning}
\bibfield{author}{\bibinfo{person}{Alec Radford}, \bibinfo{person}{Jong~Wook Kim}, \bibinfo{person}{Chris Hallacy}, \bibinfo{person}{Aditya Ramesh}, \bibinfo{person}{Gabriel Goh}, \bibinfo{person}{Sandhini Agarwal}, \bibinfo{person}{Girish Sastry}, \bibinfo{person}{Amanda Askell}, \bibinfo{person}{Pamela Mishkin}, \bibinfo{person}{Jack Clark}, {et~al\mbox{.}}} \bibinfo{year}{2021}\natexlab{}.
\newblock \showarticletitle{Learning transferable visual models from natural language supervision}. In \bibinfo{booktitle}{\emph{ICML}}. PmLR, \bibinfo{pages}{8748--8763}.
\newblock


\bibitem[Raj et~al\mbox{.}(2023)]%
        {raj2023dreambooth3d}
\bibfield{author}{\bibinfo{person}{Amit Raj}, \bibinfo{person}{Srinivas Kaza}, \bibinfo{person}{Ben Poole}, \bibinfo{person}{Michael Niemeyer}, \bibinfo{person}{Nataniel Ruiz}, \bibinfo{person}{Ben Mildenhall}, \bibinfo{person}{Shiran Zada}, \bibinfo{person}{Kfir Aberman}, \bibinfo{person}{Michael Rubinstein}, \bibinfo{person}{Jonathan Barron}, {et~al\mbox{.}}} \bibinfo{year}{2023}\natexlab{}.
\newblock \showarticletitle{Dreambooth3d: Subject-driven text-to-3d generation}. In \bibinfo{booktitle}{\emph{CVPR}}. \bibinfo{pages}{2349--2359}.
\newblock


\bibitem[Ruiz et~al\mbox{.}(2023)]%
        {ruiz2023dreambooth}
\bibfield{author}{\bibinfo{person}{Nataniel Ruiz}, \bibinfo{person}{Yuanzhen Li}, \bibinfo{person}{Varun Jampani}, \bibinfo{person}{Yael Pritch}, \bibinfo{person}{Michael Rubinstein}, {and} \bibinfo{person}{Kfir Aberman}.} \bibinfo{year}{2023}\natexlab{}.
\newblock \showarticletitle{Dreambooth: Fine tuning text-to-image diffusion models for subject-driven generation}. In \bibinfo{booktitle}{\emph{CVPR}}. \bibinfo{pages}{22500--22510}.
\newblock


\bibitem[Salemi et~al\mbox{.}(2024)]%
        {salemi-etal-2024-lamp}
\bibfield{author}{\bibinfo{person}{Alireza Salemi}, \bibinfo{person}{Sheshera Mysore}, \bibinfo{person}{Michael Bendersky}, {and} \bibinfo{person}{Hamed Zamani}.} \bibinfo{year}{2024}\natexlab{}.
\newblock \showarticletitle{{L}a{MP}: When Large Language Models Meet Personalization}. In \bibinfo{booktitle}{\emph{ACL}}. \bibinfo{publisher}{ACL}.
\newblock


\bibitem[Shen et~al\mbox{.}(2024)]%
        {shen2024pmg}
\bibfield{author}{\bibinfo{person}{Xiaoteng Shen}, \bibinfo{person}{Rui Zhang}, \bibinfo{person}{Xiaoyan Zhao}, \bibinfo{person}{Jieming Zhu}, {and} \bibinfo{person}{Xi Xiao}.} \bibinfo{year}{2024}\natexlab{}.
\newblock \showarticletitle{Pmg: Personalized multimodal generation with large language models}. In \bibinfo{booktitle}{\emph{WWW}}. \bibinfo{pages}{3833--3843}.
\newblock


\bibitem[Simonyan and Zisserman(2014)]%
        {simonyan2014very}
\bibfield{author}{\bibinfo{person}{Karen Simonyan} {and} \bibinfo{person}{Andrew Zisserman}.} \bibinfo{year}{2014}\natexlab{}.
\newblock \showarticletitle{Very deep convolutional networks for large-scale image recognition}.
\newblock \bibinfo{journal}{\emph{arXiv:1409.1556}} (\bibinfo{year}{2014}).
\newblock


\bibitem[Somepalli et~al\mbox{.}(2024)]%
        {somepalli2024measuring}
\bibfield{author}{\bibinfo{person}{Gowthami Somepalli}, \bibinfo{person}{Anubhav Gupta}, \bibinfo{person}{Kamal Gupta}, \bibinfo{person}{Shramay Palta}, \bibinfo{person}{Micah Goldblum}, \bibinfo{person}{Jonas Geiping}, \bibinfo{person}{Abhinav Shrivastava}, {and} \bibinfo{person}{Tom Goldstein}.} \bibinfo{year}{2024}\natexlab{}.
\newblock \showarticletitle{Measuring style similarity in diffusion models}.
\newblock \bibinfo{journal}{\emph{arXiv:2404.01292}} (\bibinfo{year}{2024}).
\newblock


\bibitem[Team(2024)]%
        {team2024chameleon}
\bibfield{author}{\bibinfo{person}{Chameleon Team}.} \bibinfo{year}{2024}\natexlab{}.
\newblock \showarticletitle{Chameleon: Mixed-modal early-fusion foundation models}.
\newblock \bibinfo{journal}{\emph{arXiv:2405.09818}} (\bibinfo{year}{2024}).
\newblock


\bibitem[Vashishtha et~al\mbox{.}(2024)]%
        {vashishtha2024chaining}
\bibfield{author}{\bibinfo{person}{Shanu Vashishtha}, \bibinfo{person}{Abhinav Prakash}, \bibinfo{person}{Lalitesh Morishetti}, \bibinfo{person}{Kaushiki Nag}, \bibinfo{person}{Yokila Arora}, \bibinfo{person}{Sushant Kumar}, {and} \bibinfo{person}{Kannan Achan}.} \bibinfo{year}{2024}\natexlab{}.
\newblock \showarticletitle{Chaining text-to-image and large language model: A novel approach for generating personalized e-commerce banners}. In \bibinfo{booktitle}{\emph{KDD}}. \bibinfo{pages}{5825--5835}.
\newblock


\bibitem[Wang et~al\mbox{.}(2024a)]%
        {wang2024disco}
\bibfield{author}{\bibinfo{person}{Tan Wang}, \bibinfo{person}{Linjie Li}, \bibinfo{person}{Kevin Lin}, \bibinfo{person}{Yuanhao Zhai}, \bibinfo{person}{Chung-Ching Lin}, \bibinfo{person}{Zhengyuan Yang}, \bibinfo{person}{Hanwang Zhang}, \bibinfo{person}{Zicheng Liu}, {and} \bibinfo{person}{Lijuan Wang}.} \bibinfo{year}{2024}\natexlab{a}.
\newblock \showarticletitle{Disco: Disentangled control for realistic human dance generation}. In \bibinfo{booktitle}{\emph{CVPR}}. \bibinfo{pages}{9326--9336}.
\newblock


\bibitem[Wang et~al\mbox{.}(2024b)]%
        {wang2024g}
\bibfield{author}{\bibinfo{person}{Xianquan Wang}, \bibinfo{person}{Likang Wu}, \bibinfo{person}{Shukang Yin}, \bibinfo{person}{Zhi Li}, \bibinfo{person}{Yanjiang Chen}, \bibinfo{person}{Hufeng Hufeng}, \bibinfo{person}{Yu Su}, {and} \bibinfo{person}{Qi Liu}.} \bibinfo{year}{2024}\natexlab{b}.
\newblock \showarticletitle{I-AM-G: Interest Augmented Multimodal Generator for Item Personalization}. In \bibinfo{booktitle}{\emph{EMNLP}}. \bibinfo{pages}{21303--21317}.
\newblock


\bibitem[Wei et~al\mbox{.}(2024a)]%
        {wei2024masterweaver}
\bibfield{author}{\bibinfo{person}{Yuxiang Wei}, \bibinfo{person}{Zhilong Ji}, \bibinfo{person}{Jinfeng Bai}, \bibinfo{person}{Hongzhi Zhang}, \bibinfo{person}{Lei Zhang}, {and} \bibinfo{person}{Wangmeng Zuo}.} \bibinfo{year}{2024}\natexlab{a}.
\newblock \showarticletitle{Masterweaver: Taming editability and face identity for personalized text-to-image generation}. In \bibinfo{booktitle}{\emph{ECCV}}. Springer, \bibinfo{pages}{252--271}.
\newblock


\bibitem[Wei et~al\mbox{.}(2024b)]%
        {wei2024dreamvideo}
\bibfield{author}{\bibinfo{person}{Yujie Wei}, \bibinfo{person}{Shiwei Zhang}, \bibinfo{person}{Zhiwu Qing}, \bibinfo{person}{Hangjie Yuan}, \bibinfo{person}{Zhiheng Liu}, \bibinfo{person}{Yu Liu}, \bibinfo{person}{Yingya Zhang}, \bibinfo{person}{Jingren Zhou}, {and} \bibinfo{person}{Hongming Shan}.} \bibinfo{year}{2024}\natexlab{b}.
\newblock \showarticletitle{Dreamvideo: Composing your dream videos with customized subject and motion}. In \bibinfo{booktitle}{\emph{CVPR}}. \bibinfo{pages}{6537--6549}.
\newblock


\bibitem[Wu et~al\mbox{.}(2025)]%
        {wu2025core}
\bibfield{author}{\bibinfo{person}{Feize Wu}, \bibinfo{person}{Yun Pang}, \bibinfo{person}{Junyi Zhang}, \bibinfo{person}{Lianyu Pang}, \bibinfo{person}{Jian Yin}, \bibinfo{person}{Baoquan Zhao}, \bibinfo{person}{Qing Li}, {and} \bibinfo{person}{Xudong Mao}.} \bibinfo{year}{2025}\natexlab{}.
\newblock \showarticletitle{Core: Context-regularized text embedding learning for text-to-image personalization}. In \bibinfo{booktitle}{\emph{AAAI}}, Vol.~\bibinfo{volume}{39}. \bibinfo{pages}{8377--8385}.
\newblock


\bibitem[Wu et~al\mbox{.}(2024)]%
        {wu2024videomaker}
\bibfield{author}{\bibinfo{person}{Tao Wu}, \bibinfo{person}{Yong Zhang}, \bibinfo{person}{Xiaodong Cun}, \bibinfo{person}{Zhongang Qi}, \bibinfo{person}{Junfu Pu}, \bibinfo{person}{Huanzhang Dou}, \bibinfo{person}{Guangcong Zheng}, \bibinfo{person}{Ying Shan}, {and} \bibinfo{person}{Xi Li}.} \bibinfo{year}{2024}\natexlab{}.
\newblock \showarticletitle{Videomaker: Zero-shot customized video generation with the inherent force of video diffusion models}.
\newblock \bibinfo{journal}{\emph{arXiv:2412.19645}} (\bibinfo{year}{2024}).
\newblock


\bibitem[Xiao et~al\mbox{.}(2025)]%
        {xiao2025omnigen}
\bibfield{author}{\bibinfo{person}{Shitao Xiao}, \bibinfo{person}{Yueze Wang}, \bibinfo{person}{Junjie Zhou}, \bibinfo{person}{Huaying Yuan}, \bibinfo{person}{Xingrun Xing}, \bibinfo{person}{Ruiran Yan}, \bibinfo{person}{Chaofan Li}, \bibinfo{person}{Shuting Wang}, \bibinfo{person}{Tiejun Huang}, {and} \bibinfo{person}{Zheng Liu}.} \bibinfo{year}{2025}\natexlab{}.
\newblock \showarticletitle{Omnigen: Unified image generation}. In \bibinfo{booktitle}{\emph{CVPR}}. \bibinfo{pages}{13294--13304}.
\newblock


\bibitem[Xie et~al\mbox{.}(2025)]%
        {xie2025showo}
\bibfield{author}{\bibinfo{person}{Jinheng Xie}, \bibinfo{person}{Weijia Mao}, \bibinfo{person}{Zechen Bai}, \bibinfo{person}{David~Junhao Zhang}, \bibinfo{person}{Weihao Wang}, \bibinfo{person}{Kevin~Qinghong Lin}, \bibinfo{person}{Yuchao Gu}, \bibinfo{person}{Zhijie Chen}, \bibinfo{person}{Zhenheng Yang}, {and} \bibinfo{person}{Mike~Zheng Shou}.} \bibinfo{year}{2025}\natexlab{}.
\newblock \showarticletitle{Show-o: One Single Transformer to Unify Multimodal Understanding and Generation}. In \bibinfo{booktitle}{\emph{ICLR}}.
\newblock


\bibitem[Xu et~al\mbox{.}(2024a)]%
        {xu2024disentangled}
\bibfield{author}{\bibinfo{person}{Wenju Xu}, \bibinfo{person}{Chengjiang Long}, \bibinfo{person}{Yongwei Nie}, {and} \bibinfo{person}{Guanghui Wang}.} \bibinfo{year}{2024}\natexlab{a}.
\newblock \showarticletitle{Disentangled representation learning for controllable person image generation}.
\newblock \bibinfo{journal}{\emph{TMM}}  \bibinfo{volume}{26} (\bibinfo{year}{2024}), \bibinfo{pages}{6065--6077}.
\newblock


\bibitem[Xu et~al\mbox{.}(2024b)]%
        {xu2024diffusion}
\bibfield{author}{\bibinfo{person}{Yiyan Xu}, \bibinfo{person}{Wenjie Wang}, \bibinfo{person}{Fuli Feng}, \bibinfo{person}{Yunshan Ma}, \bibinfo{person}{Jizhi Zhang}, {and} \bibinfo{person}{Xiangnan He}.} \bibinfo{year}{2024}\natexlab{b}.
\newblock \showarticletitle{Diffusion models for generative outfit recommendation}. In \bibinfo{booktitle}{\emph{SIGIR}}. \bibinfo{pages}{1350--1359}.
\newblock


\bibitem[Xu et~al\mbox{.}(2024c)]%
        {xu2024personalized}
\bibfield{author}{\bibinfo{person}{Yiyan Xu}, \bibinfo{person}{Wenjie Wang}, \bibinfo{person}{Yang Zhang}, \bibinfo{person}{Biao Tang}, \bibinfo{person}{Peng Yan}, \bibinfo{person}{Fuli Feng}, {and} \bibinfo{person}{Xiangnan He}.} \bibinfo{year}{2024}\natexlab{c}.
\newblock \showarticletitle{Personalized Image Generation with Large Multimodal Models}.
\newblock \bibinfo{journal}{\emph{arXiv:2410.14170}} (\bibinfo{year}{2024}).
\newblock


\bibitem[Xu et~al\mbox{.}(2024d)]%
        {xu2024sgdm}
\bibfield{author}{\bibinfo{person}{Yifei Xu}, \bibinfo{person}{Xiaolong Xu}, \bibinfo{person}{Honghao Gao}, {and} \bibinfo{person}{Fu Xiao}.} \bibinfo{year}{2024}\natexlab{d}.
\newblock \showarticletitle{Sgdm: an adaptive style-guided diffusion model for personalized text to image generation}.
\newblock \bibinfo{journal}{\emph{TMM}} (\bibinfo{year}{2024}).
\newblock


\bibitem[Xu et~al\mbox{.}(2025)]%
        {xu2025personalized}
\bibfield{author}{\bibinfo{person}{Yiyan Xu}, \bibinfo{person}{Jinghao Zhang}, \bibinfo{person}{Alireza Salemi}, \bibinfo{person}{Xinting Hu}, \bibinfo{person}{Wenjie Wang}, \bibinfo{person}{Fuli Feng}, \bibinfo{person}{Hamed Zamani}, \bibinfo{person}{Xiangnan He}, {and} \bibinfo{person}{Tat-Seng Chua}.} \bibinfo{year}{2025}\natexlab{}.
\newblock \showarticletitle{Personalized Generation In Large Model Era: A Survey}.
\newblock \bibinfo{journal}{\emph{arXiv:2503.02614}} (\bibinfo{year}{2025}).
\newblock


\bibitem[Yang et~al\mbox{.}(2024a)]%
        {yang2024qwen2}
\bibfield{author}{\bibinfo{person}{An Yang}, \bibinfo{person}{Baosong Yang}, \bibinfo{person}{Beichen Zhang}, \bibinfo{person}{Binyuan Hui}, \bibinfo{person}{Bo Zheng}, \bibinfo{person}{Bowen Yu}, \bibinfo{person}{Chengyuan Li}, \bibinfo{person}{Dayiheng Liu}, \bibinfo{person}{Fei Huang}, \bibinfo{person}{Haoran Wei}, {et~al\mbox{.}}} \bibinfo{year}{2024}\natexlab{a}.
\newblock \showarticletitle{Qwen2.5 technical report}.
\newblock \bibinfo{journal}{\emph{arXiv:2412.15115}} (\bibinfo{year}{2024}).
\newblock


\bibitem[Yang et~al\mbox{.}(2024b)]%
        {yang2024new}
\bibfield{author}{\bibinfo{person}{Hao Yang}, \bibinfo{person}{Jianxin Yuan}, \bibinfo{person}{Shuai Yang}, \bibinfo{person}{Linhe Xu}, \bibinfo{person}{Shuo Yuan}, {and} \bibinfo{person}{Yifan Zeng}.} \bibinfo{year}{2024}\natexlab{b}.
\newblock \showarticletitle{A New Creative Generation Pipeline for Click-Through Rate with Stable Diffusion Model}. In \bibinfo{booktitle}{\emph{WWW}}. \bibinfo{pages}{180--189}.
\newblock


\bibitem[Zhang et~al\mbox{.}(2018)]%
        {zhang2018unreasonable}
\bibfield{author}{\bibinfo{person}{Richard Zhang}, \bibinfo{person}{Phillip Isola}, \bibinfo{person}{Alexei~A Efros}, \bibinfo{person}{Eli Shechtman}, {and} \bibinfo{person}{Oliver Wang}.} \bibinfo{year}{2018}\natexlab{}.
\newblock \showarticletitle{The unreasonable effectiveness of deep features as a perceptual metric}. In \bibinfo{booktitle}{\emph{CVPR}}. \bibinfo{pages}{586--595}.
\newblock


\bibitem[Zhou et~al\mbox{.}(2025)]%
        {zhou2025transfusion}
\bibfield{author}{\bibinfo{person}{Chunting Zhou}, \bibinfo{person}{LILI YU}, \bibinfo{person}{Arun Babu}, \bibinfo{person}{Kushal Tirumala}, \bibinfo{person}{Michihiro Yasunaga}, \bibinfo{person}{Leonid Shamis}, \bibinfo{person}{Jacob Kahn}, \bibinfo{person}{Xuezhe Ma}, \bibinfo{person}{Luke Zettlemoyer}, {and} \bibinfo{person}{Omer Levy}.} \bibinfo{year}{2025}\natexlab{}.
\newblock \showarticletitle{Transfusion: Predict the Next Token and Diffuse Images with One Multi-Modal Model}. In \bibinfo{booktitle}{\emph{ICLR}}.
\newblock


\bibitem[Zhu et~al\mbox{.}(2024)]%
        {zhu2024minigpt}
\bibfield{author}{\bibinfo{person}{Deyao Zhu}, \bibinfo{person}{Jun Chen}, \bibinfo{person}{Xiaoqian Shen}, \bibinfo{person}{Xiang Li}, {and} \bibinfo{person}{Mohamed Elhoseiny}.} \bibinfo{year}{2024}\natexlab{}.
\newblock \showarticletitle{Mini{GPT}-4: Enhancing Vision-Language Understanding with Advanced Large Language Models}. In \bibinfo{booktitle}{\emph{ICLR}}.
\newblock


\end{thebibliography}
}

% \newpage
\appendix

\section{Appendix}
\subsection{Examples of Contrastive Samples}
\label{sec:contrastive_example}
\begin{figure}[H]
\setlength{\abovecaptionskip}{0.cm}
\centering
\includegraphics[scale=0.44]{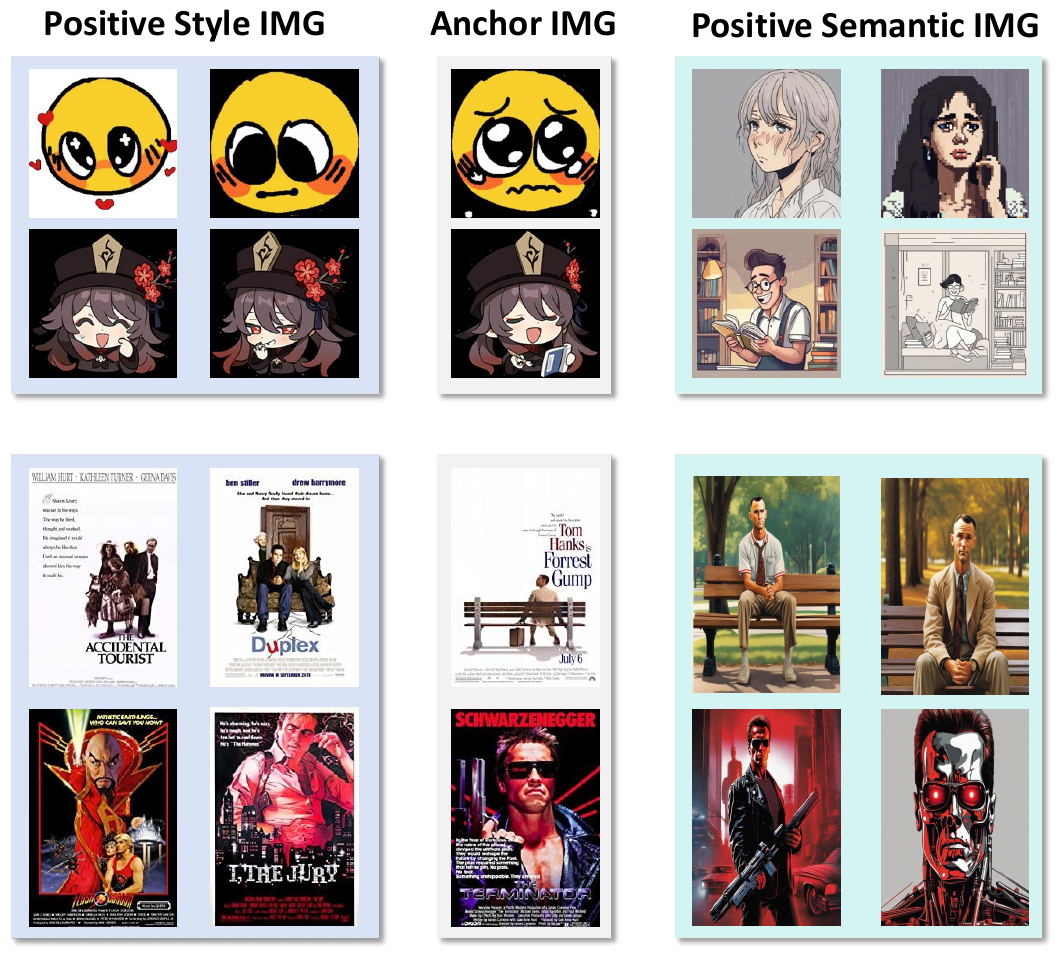}
\caption{Examples of contrastive samples.} 
\label{fig:triplet_examples}
\end{figure}

In Figure~\ref{fig:triplet_examples}, we present several contrastive samples for both sticker and movie poster scenarios, where positive style images share similar style with the anchor but differ in semantics, while positive semantic images share the same semantics but differ in style.

\subsection{Examples of DRC}
\label{sec:cases}
\begin{figure}[H]
\setlength{\abovecaptionskip}{0.1cm}
\centering
\includegraphics[scale=0.4]{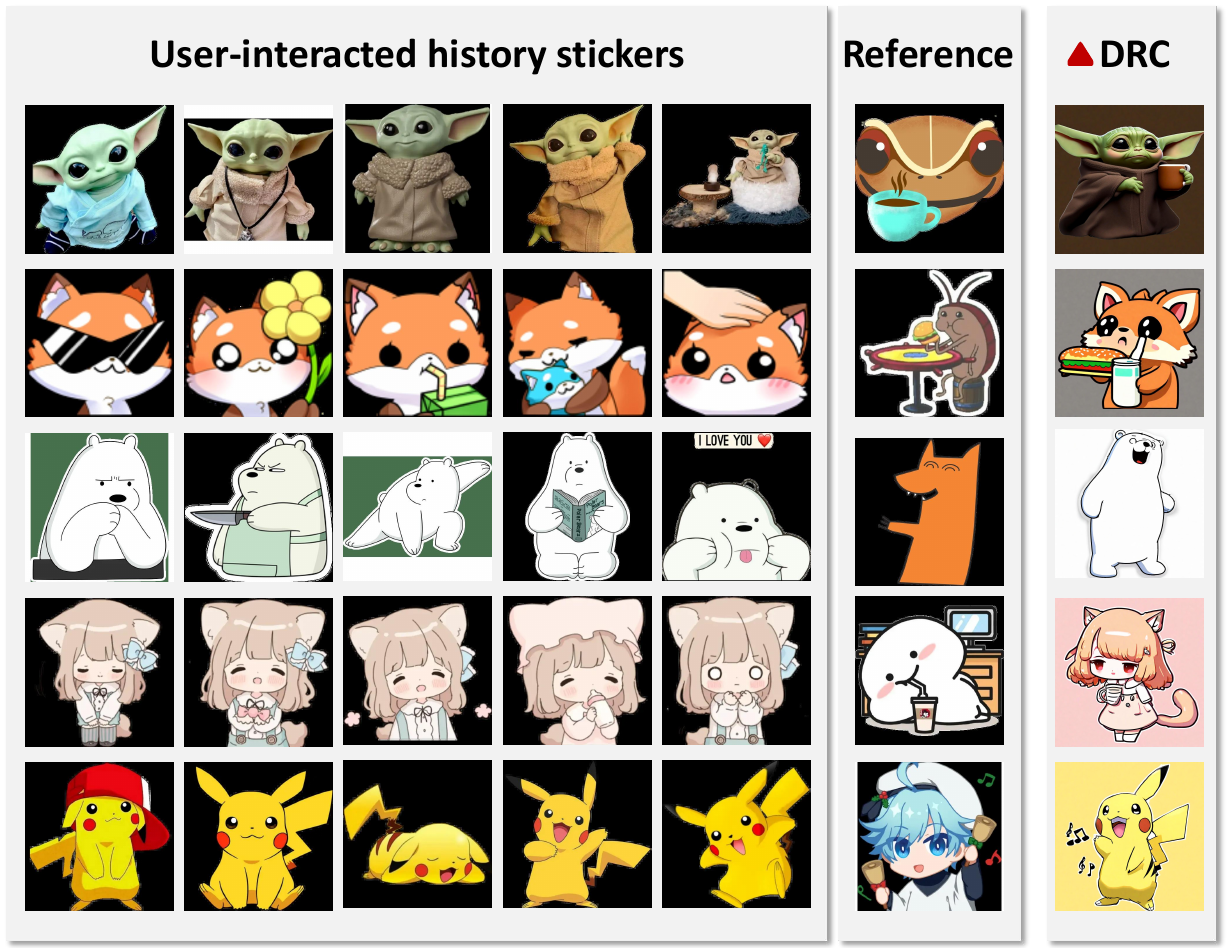}
\caption{Stickers generated by DRC.} 
\label{fig:sticker_ex}
\end{figure}

As shown in Figure~\ref{fig:sticker_ex}, DRC effectively captures the character identity and overall visual style from the history images. Meanwhile, it accurately reflects the character's pose and the semantic intent conveyed by the reference image.

\subsection{Computational Costs}
\label{sec:cost}
We present a comparison of training and inference costs between DRC and Pigeon in Table~\ref{tab:cost}. Although DRC introduces an additional disentanglement stage (Stage-1), resulting in a slightly longer total training time compared to Pigeon, it is more memory-efficient during training and achieves significantly faster inference. Overall, we believe this trade-off is favorable, particularly given that inference efficiency is often the key bottleneck in real-world deployment. More importantly, DRC effectively mitigates the guidance collapse issue, as demonstrated in the experimental results in Section~\ref{sec:experiment}.

\begin{table}[H]
\setlength{\abovecaptionskip}{0cm}
\setlength{\belowcaptionskip}{0cm}
\caption{Comparison of computational costs.}
\label{tab:cost}
\begin{tabular}{ll|ll}
\hline
                                      &            & \textbf{DRC} & \textbf{Pigeon} \\ \hline
\multicolumn{2}{l|}{\textbf{GPU Memory}}           & 55G          & 60G             \\ \hline
\multirow{2}{*}{\textbf{Training}}    & Stage-1    & 15.5h        & -               \\ \cline{2-4} 
                                      & Stage-2    & 7.5h         & 20h             \\ \hline
\multicolumn{2}{l|}{\textbf{Inference per Sample}} & 7s           & 16s             \\ \hline
\end{tabular}
\end{table}

\subsection{Human Evaluation}
\label{sec:user_study}
We conduct human evaluation on Amazon MTurk\footnote{https://www.mturk.com/.}, comparing DRC with the most competitive baseline Pigeon, using binary-choice tests on the sticker scenario with 50 cases. We evaluate both style and semantic alignment separately, recruiting 50 participants for each evaluation. DRC is consistently preferred over Pigeon in style alignment, with $74\%\pm5.4\%$ of participants favoring it, where ``$\pm$'' denotes the 95\% confidence interval. For semantic alignment, DRC achieves comparable performance ($51\%\pm3.8\%$). These findings are consistent with our quantitative analysis, further highlighting DRC's superiority in addressing the guidance collapse issue.

\end{document}